\documentclass[journal]{IEEEtran}
\usepackage{cite}
\usepackage{hyperref}

\ifCLASSINFOpdf
   \usepackage[pdftex]{graphicx}
\else
   \usepackage[dvips]{graphicx}
\fi
\usepackage{makecell}
\usepackage{booktabs}
\usepackage{xcolor}
\usepackage{multirow}

\usepackage{amsmath,amssymb}
\DeclareMathOperator{\Tr}{Tr}
\usepackage{pifont}
\newcommand{\cmark}{\ding{51}}
\newcommand{\xmark}{\ding{55}}

\usepackage{url}

\begin{document}
%
\title{Human Pose and Shape Estimation from Single Polarization Images}
%
%
%

\author{Shihao~Zou,~\IEEEmembership{Member,~IEEE},
        Xinxin~Zuo,
        Sen~Wang,
        Yiming~Qian, 
        Chuan~Guo,
        and~Li~Cheng,~\IEEEmembership{Senior~Member,~IEEE}
\thanks{S. Zou, X. Zuo, S. Wang and C. Guo are with the Department
of Electrical and Computer Engineering, University of Alberta, AB, Canada. (E-mail: szou2@ualberta.ca, xzuo@ualberta.ca, sen9@ualberta.ca, cguo2@ualberta.ca).}
\thanks{Y. Qian is with the Department of Computer Science, University of Manitoba, MB , Canada. (E-mail: yiming.qian@umanitoba.ca)}
\thanks{L. Cheng is the Department of Electrical and Computer Engineering, University of Alberta, AB, Canada. (E-mail: lcheng5@ualberta.ca)}
\thanks{Manuscript received January 05 2022; accepted March 18 2022. Li Cheng is the corresponding author for this paper.}}

%
%

\markboth{Journal of \LaTeX\ Class Files,~Vol.~14, No.~8, August~2015}%
{Shell \MakeLowercase{\textit{et al.}}: Bare Demo of IEEEtran.cls for IEEE Journals}
%



\maketitle

\begin{abstract}
This paper focuses on a new problem of estimating human pose and shape from single polarization images. Polarization camera is known to be able to capture the polarization of reflected lights that preserves rich geometric cues of an object surface. Inspired by the recent applications in  surface normal reconstruction from polarization images, in this paper, we attempt to estimate human pose and shape from single polarization images by leveraging the polarization-induced geometric cues. A dedicated two-stage pipeline is proposed: given a single polarization image, stage one (Polar2Normal) focuses on the fine detailed human body surface normal estimation; stage two (Polar2Shape) then reconstructs clothed human shape from the polarization image and the estimated surface normal. To empirically validate our approach, a dedicated dataset (PHSPD) is constructed, consisting of over 500K frames with accurate pose and parametric shape annotations. Empirical evaluations on this real-world dataset as well as a synthetic dataset, SURREAL, demonstrate the effectiveness of our approach. It suggests polarization camera as a promising alternative to the more conventional RGB camera for human pose and shape estimation.

\end{abstract}


%
\IEEEpeerreviewmaketitle

\section{Introduction}

A critical computer vision problem is to predict 3D human poses, i.e. 3D body joint locations, from single images. In recent years, rapid progress is made in 3D human pose prediction from \textit{RGB} images~\cite{ning2017knowledge,kamel2020hybrid,zhang2021single,park20163d,zhao2017simple,nie2017monocular,wang2018drpose3d,fang2018learning,habibie2019wild,wang20193d}. Moreover, fueled by the development in parametric human body shape modelling, such as SCAPE~\cite{anguelov2005scape} and SMPL~\cite{loper2015smpl}, it becomes feasible to estimate human body shapes from single RGB image, as is evidenced by a number of end-to-end deep learning methods~\cite{zhao20183,lu2020parametric,zuo2020sparsefusion,balan2007detailed,dibra2016hs,bogo2016keep,lassner2017unite,kanazawa2018end,pavlakos2018learning,kanazawa2019learning,kocabas2020vibe,zheng2021pamir,zhu2021detailed}. 
On the other hand, the problems of 3D human pose and shape estimation from single RGB images are still far from being solved. This is mainly due to the inherently lack of 3D cues in an RGB image. 
Furthermore, as the human shape models (e.g. SMPL) are usually learned from large sets of scanned human naked bodies, they are often lacking in clothing details.

The above observation inspires us to investigate in this paper a new imaging modality, polarization images. That is, we consider the problem of estimating human pose and shape from a single polarization image. Polarization camera is realised upon a basic physics principle: a light ray reflected from an object is usually polarized. The polarized signal thus carries sufficient geometric cues of object surface details to reliably infer its surface normal~\cite{yang2018polarimetric,ba2019physics}. 
It is worth mentioning the biological fact that light polarization could be directly perceived by some species of bees, ants, and shrimps for purposes such as 3D navigation~\cite{wehner2006significance,daly2016dynamic}. 
Motivated by the physical and biological facts, we propose a dedicated two-stage approach for human pose and shape estimation by integrating the geometric cues from the input polarization images.

\begin{figure}
    \centering
    \includegraphics[width=\columnwidth]{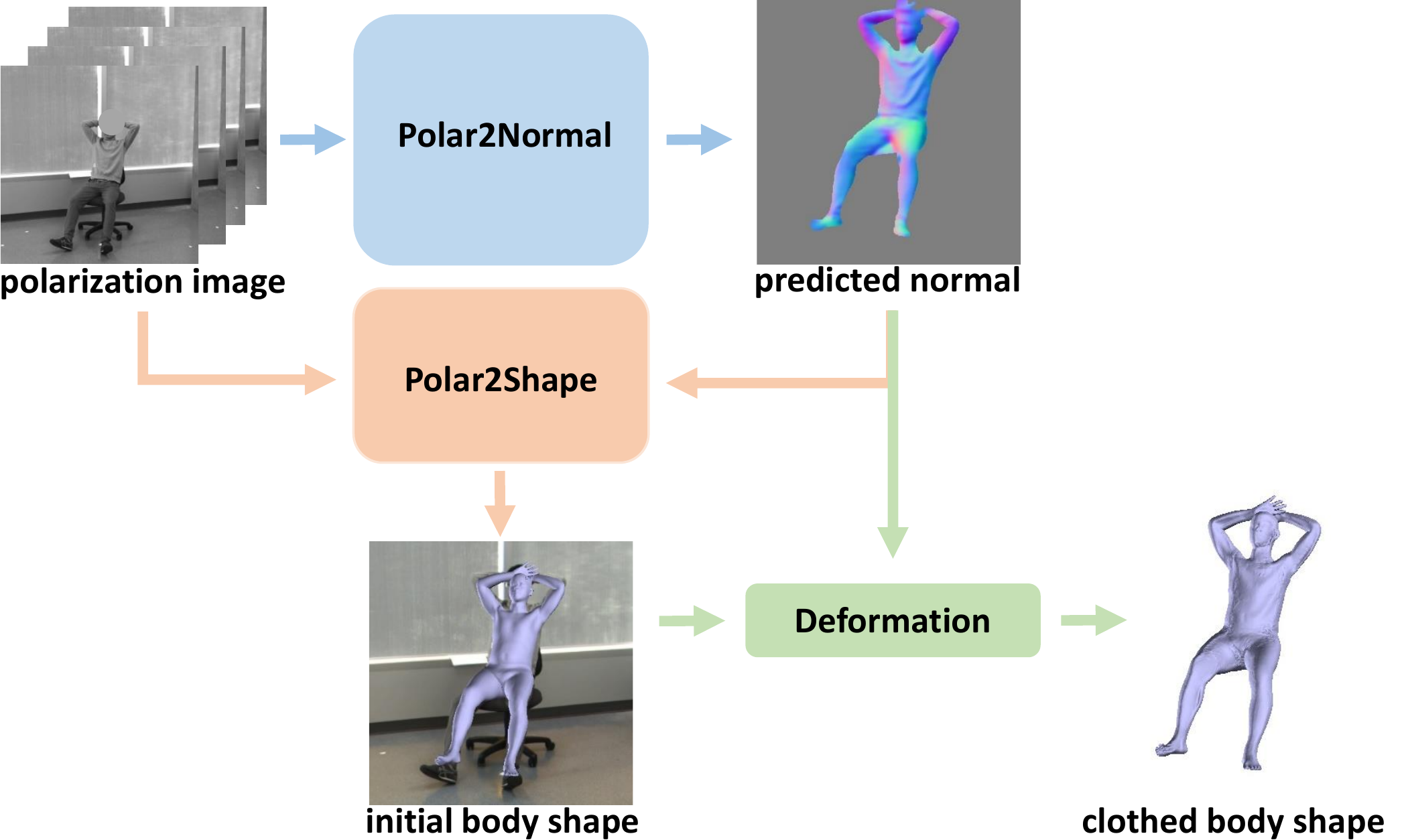}
    \caption{An overview of our HumanSfP approach that consists of two main stages: stage one, \emph{Polar2Normal}, infers surface normal from single polarization images; stage two, \emph{Polar2Shape}, reconstructs clothed human body shape by deforming the initial body shape guided by the predicted surface normal, where the initial shape is parameterized by SMPL model, estimated from the raw polarization image and the surface normal.}
    \label{fig:overview}
\end{figure}

As shown in Fig.~\ref{fig:overview}, our approach, also called HumanSfP, contains two main stages. Stage one, Polar2Normal, concentrates on predicting accurate surface normal maps from single polarization images by exploiting the associated physics laws as priors. It is then fed into stage two, Polar2Shape, in reconstructing a clothed human shape guided by the obtained surface normal and an initial shape estimated by SMPL.

Different from previous efforts in estimating detailed depth map~\cite{tang2019neural,zhu2021detailed}, our stage one focuses on surface normal estimation from a polarization image. By explicitly integrating the underlying physical principles, it gives rise to more reliable estimation. To achieve this goal, two main challenges need to be addressed, namely $\pi$-ambiguity of the azimuth angle and the possible environmental noise in practical applications. Based on the physical laws of polarization, two ambiguous normal maps $\mathbf{m}_1$ and $\mathbf{m}_2$ (Sec. \ref{sec:normal-estimation}) can be directly calculated from a polarization image. It is often reasonable to assume light reflected by human clothes is dominated by diffused reflection. Then, different from~\cite{ba2019physics}, we consider a two-branch strategy: one branch is employed to categorize each pixel into three categories: the two ambiguous normal maps and background; the second branch is to infer a coarse surface normal map. A fused normal map is produced by incorporating the classification results as well as the two ambiguous normal maps. 
Finally, as is revealed by~\cite{zou2020detailed}, the fused normal map output may still be noisy, owing to environmental noise and the digital quantization of the polarization camera. We instead work with the normal residual calculated as the difference between the coarse and the fused normal map, which is used to refine the coarse normal map, and to produce our final normal estimation. 

Based on the raw polarization image and final surface normal output of stage one, stage two concerns the estimation of 3D human pose and the reconstruction of its clothed shape. It starts from estimating an initial parametric shape, i.e. SMPL shape, as 3D pose, which is then deformed by leveraging the geometric details from the surface normal, to reconstruct the final clothed human shape. Different from previous works~\cite{zou2020detailed,kanazawa2018end,kanazawa2019learning}, geodesic distance is employed in SMPL shape estimation, since the pose representation of SMPL is naturally in a product space of $\text{SO}(3)$, a classical example of Lie group. Our two-stage pipeline is empirically shown to be capable of inferring detailed surface normal faithfully, and estimating human pose and clothed body shape accurately.

To summarize, there are three main contributions in our work. (i) A new problem, namely human pose and shape estimation from single polarization images, is proposed. A dedicated deep learning approach, HumanSfP, is proposed, where the detail-preserving surface normal maps are obtained following the physical laws of light polarization, and are shown to estimate more accurate pose and body shape. (ii) In tackling this new problem, a dedicated Polarization Human Shape and Pose Dataset, PHSPD, has been created. It now consists of $\sim$527K frames and their corresponding pose and shape annotations. Overall there are 21 different subjects performing 31 unique actions, and $\sim$9.5 hours of videos are recorded in total. The dataset and our code are publicly available\footnote{ \href{https://github.com/JimmyZou/PolarHumanPoseShape}{https://github.com/JimmyZou/PolarHumanPoseShape}}. (iii) Empirical evaluations on a synthetic dataset, SURREAL dataset, as well as our real-world dataset, PHSPD dataset, demonstrate the effectiveness and applicability of our approach. Our work showcases that estimate 3D human poses and shapes, 2D polarization camera could be a viable alternative to conventional RGB cameras.

Our preliminary work is published in~\cite{zou2020detailed}. This paper extends our preceding effort in a number of aspects.  
\begin{itemize}
    \item Extending from our PHSPDv1~\cite{zou2020detailed} dataset of 287K frames, 240K additional frames with \emph{annotated} poses and parametric shapes are contained in PHSPDv2. Each new polarization frame in PHSPDv2 is synchronized by five RGBD images of distinct views and one event camera; this is compared with only three RGBD cameras in PHSPDv1. Besides, PHSPDv2 is more versatile with 15 subjects performing 21 different actions, comparing to PHSPDv1 of 12 subjects performing 18 different actions. Overall, the extended PHSPD dataset is the largest to date against other 3D human pose estimation datasets~\cite{human36m,mehta2017monocular}, and its multi-modality property renders its great potential in facilitating existing and new research directions.
    \item A new dedicated residual update module is designed for surface normal estimation from a polarization image. Empirical evaluations demonstrate its superior performance, achieved by better leveraging geometric cues from polarization images. Moreover, during 3D pose estimation, Lie group-based geodesic distance is introduced instead of the popular Euclidean distance in~\cite{zou2020detailed,kanazawa2018end,kocabas2020vibe}. This is practically shown to well-measure the predicted and the target poses. This we attribute to the physical laws of human kinematics that takes place in the product space of $\text{SO}(3)$, which is a Lie group. 
    \item Compared to the previous work~\cite{zou2020detailed}, our newly proposed approach consistently improves the performance of surface normal estimation, pose estimation and clothed shape reconstruction on the related benchmarks. Further, our proposed model for normal estimation also achieves the SOTA results for the task of Shape from Polarization (SfP). Together they demonstrate the effectiveness of our new approach.
\end{itemize}

\section{Related Work}
\paragraph{Shape from polarization (SfP)} It is a related task that focuses on inferring shape (normally represented as surface normal) from a polarization image, where each channel captures the polarimetric information of reflected light under a linear polarizer with a different angle. There are two main issues involved in SfP: angle ambiguity and the discrimination of specular and diffuse reflection. Previous efforts are mainly physics-based that rely on additional information or assumptions to elucidate the possible ambiguities, such as smooth object surfaces \cite{atkinson2006recovery}, coarse depth map~\cite{kadambi2017depth,yang2018polarimetric} and multi-view geometric constraint \cite{chen2018polarimetric,cui2017polarimetric}. The first deep learning based method is conceived in~\cite{ba2019physics} that integrates physical priors (ambiguous normal maps) with deep models in estimating normal maps. It has shown that deep model can learn to leverage the angle ambiguity and environmental noise. A following-up work~\cite{zou2020detailed} advocates to first classify each pixel into different types of ambiguous angle and obtain a fused normal map, which is shown to extract more explicit geometric cues from polarization images.

\paragraph{Human Pose Estimation from Single Images} In the past few years, 3D human pose estimation from single images, mainly based on RGB or depth images, has been extensively studied. Many early efforts~\cite{zhou2016sparseness,akhter2015pose,wang2014robust}
utilize dictionary-based learning strategies to capture prior knowledge from large motion-capture dataset. Recent efforts focus on end-to-end deep learning based methods, including CNNs~\cite{park20163d,li2015maximum} and Graph CNNs~\cite{ci2019optimizing,cai2019exploiting} to estimate 3D human poses. In particular, a common framework has been adopted by a number of recent works~\cite{tome2017lifting,martinez2017simple,zhao2017simple,zhou2017towards,yang20183d,pavlakos2018ordinal,wang20193d,wandt2019repnet,habibie2019wild}, which first infer 2D poses (either 2D joint positions or heatmaps), which are then lifted to 3D. Self-supervised learning \cite{wang20193d,habibie2019wild} and adversarial learning~\cite{yang20183d,wandt2019repnet} are also considered to exploit the benefits of additional re-projection or adversarial constraints.

\paragraph{Human Shape Estimation from Single Images} Going beyond pose estimation, the availability of parametric human shape models such as SMPL model~\cite{loper2015smpl}, has fueled growing attentions in single image-based human shape estimation. SMPL is a statistical low-dimensional representation of human shape, realized by principal component analysis of empirical body shapes of naked and minimally dressed humans. 
Early efforts emphasize on optimization-based methods to fit SMPL model to point cloud or annotated pose \& human silhouette~\cite{bogo2016keep,balan2007detailed,dibra2017human,dibra2016hs}. Recent deep learning based methods~\cite{kanazawa2018end,varol2017learning,omran2018neural} instead learn to predict SMPL parameters under various constraints such as 2D/3D pose, silhouette, and adversarial examples. Human body pixel-to-surface correspondence map is considered in~\cite{xu2019denserac} for parametric shape estimation. In~\cite{kolotouros2019learning}, optimization and regression are integrated to form a self-improvement loop. Temporal information is also investigated in~\cite{kocabas2020vibe} that estimates human shape in a video. Point cloud is also explored in parametric human shape estimation~\cite{jiang2019skeleton}. Rather than RGB images, our work works with single polarization images. Moreover, our aim is to estimate human shapes with clothing details, which goes beyond the capability of parametric SMPL human shapes for naked and minimally dressed humans.

In terms of human shape reconstruction, volume-based methods~\cite{varol2018bodynet,zheng2019deephuman,saito2019pifu,zheng2021pamir,yang2021s3} are popular in reconstructing detailed body shapes. They unfortunately suffer from the limitation of computation scalability and lack of reliable 3D cues. The most recent work, PaMIR~\cite{zheng2021pamir}, combines the parametric body model with the free-form deep implicit function to reconstruct human shape. Unlike surface normal, the implicit field cannot provide explicit cues of human body and leads to in-accurate reconstructed shapes. Saito et al.~\cite{saito2019pifu,saito2020pifuhd} introduce a pixel-aligned implicit surface function to encode detailed body surface. However, these two works are not able to handle complex poses, partly owing to the lack of complex poses in their training set. A closely related work is~\cite{zhu2021detailed} that considers a hierarchical framework to incorporate robust parametric shape estimation and flexible 3D shape deformation. It however employs a network trained on additional small dataset to infer shading information, which are inherently unreliable given the lack of ground-truth information of surface normal, albedo and environmental lighting. On the contrary, our method infers clothed shape with the reliable normal map estimated from the polarization image. Another related work is~\cite{tang2019neural} that iteratively integrates rough depth map and estimated surface normal for improved clothes details.

\paragraph{Human Pose and Shape Dataset} A number of human pose and shape datasets have been released in recent years, including MPII~\cite{akhter2015pose}, MS COCO~\cite{lin2014microsoft} and PoseTrack~\cite{andriluka2018posetrack}, which provide in-the-wild RGB images and 2D pose annotations. Human3.6M~\cite{human36m}, MPI-INF-3DHP~\cite{mehta2017monocular} and 3DPW~\cite{von2018recovering} are three benchmark datasets with 3D annotations for human pose estimation from RGB images. There are also human shape datasets~\cite{zheng2019deephuman,tang2019neural,saito2019pifu} that mainly consist of RGBD images and the corresponding human shape annotations.
Our PHSPD dataset is the first such dataset based on polarization images.

\section{Preliminary Backgrounds}
\subsection{Polarization Image Formation} 
The reflected light from a surface mainly includes three components~\cite{cui2017polarimetric}, the polarized specular reflection, the polarized diffuse reflection, and the unpolarized diffuse reflection. 
A polarization camera is equipped with an array of linear polarizers mounted right on top of its CMOS imager, in place of the RGB Bayer filters. During the imaging process, a pixel's intensity typically varies sinusoidally with the angle of the polarizer~\cite{yang2018polarimetric}.
In this work, we assume that the light reflected off human clothes is dominated by polarized diffuse reflection and unpolarized diffuse reflection. For a specific polarizer angle $\phi_{\mathrm{pol}}$, the illumination intensity at a pixel with dominant diffuse reflection is
\begin{equation}
    \label{eq:polar}
    \mathrm{I}(\phi_{\mathrm{pol}})=\frac{\mathrm{I}_{\max} + \mathrm{I}_{\min}}{2} + \frac{\mathrm{I}_{\max} - \mathrm{I}_{\min}}{2} \cos \big(2(\phi_{\mathrm{pol}}-\varphi)\big).
\end{equation}
Here $\varphi$ is the azimuth angle of surface normal, $\mathrm{I}_{\max}$ and $\mathrm{I}_{\min}$ are the upper and lower bounds of the illumination intensity. $\mathrm{I}_{\max}$ and $\mathrm{I}_{\min}$ are mainly determined by the unpolarized diffuse reflection, and the sinusoidal variation is mainly determined by the polarized diffuse reflection. 
If the intensity images $\mathrm{I}(\phi_{\mathrm{pol}})$ under three or more polarizer angles can be obtained, such as $\mathrm{I}(0^{\circ})$, $\mathrm{I}(45^{\circ})$ and $\mathrm{I}(90^{\circ})$, $\varphi$ can be solved in closed form. Note that there is $\pi$-ambiguity in the azimuth angle $\varphi$ in Eq.\eqref{eq:polar}, which means that $\varphi$ and $\pi+\varphi$ will result in the same illumination intensity of the pixel. As for the zenith angle $\theta$, it is related to the degree of polarization $\rho$, 
\begin{equation}
    \label{eq:degree-of-polarization}
    \rho = \frac{\mathrm{I}_{\max} - \mathrm{I}_{\min}}{\mathrm{I}_{\max} + \mathrm{I}_{\min}}.
\end{equation}
According to \cite{atkinson2006recovery}, when diffuse reflection dominates, the degree of polarization $\rho$ becomes a function of the zenith angle $\theta$ and the refractive index $n$,
\begin{equation}
    \label{eq:zenith-angle}
    \rho = \frac{(n-\frac{1}{n})^2\sin^2\theta}{2+2n^2-(n+\frac{1}{n})^2\sin^2\theta+4\cos\theta\sqrt{n^2-\sin^2\theta}}.
\end{equation}
In this paper, we assume the refractive index $n=1.5$ since the material of human clothes is mainly cotton or nylon. The solution of $\theta$ in Eq.~\eqref{eq:zenith-angle} is thus a close-form expression of $n$ and $\rho$.

\begin{figure*}
    \centering
    \includegraphics[width=0.95\textwidth]{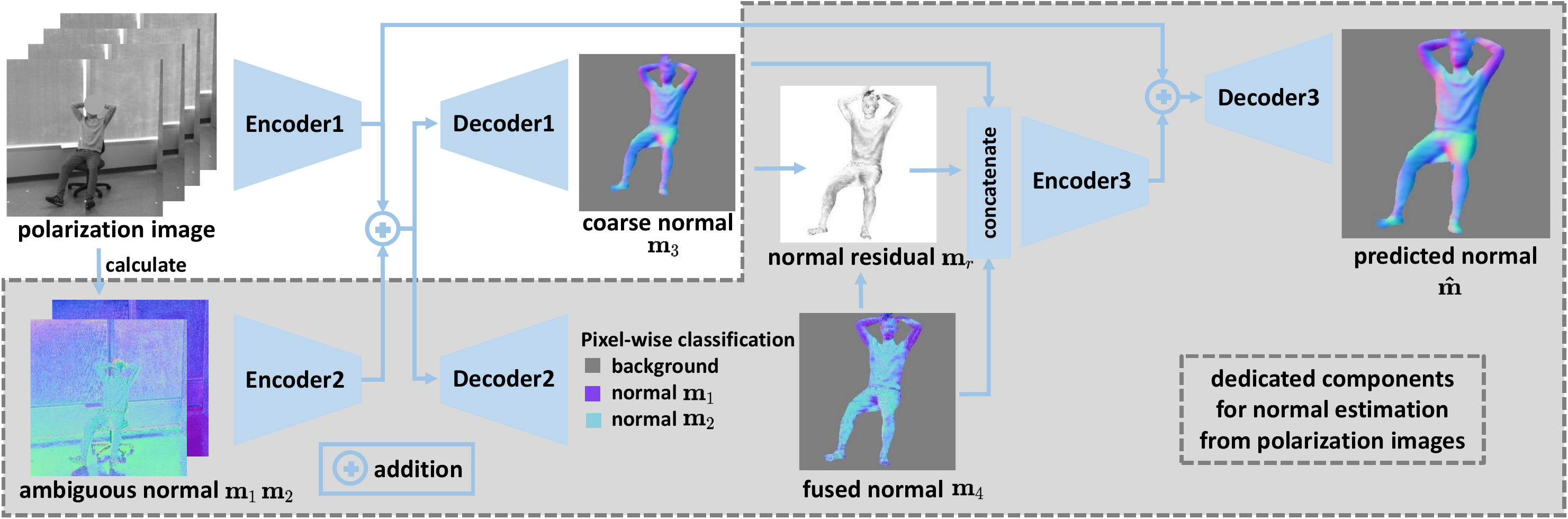}
    \caption{Stage one: our Polar2Normal pipeline for surface normal estimation from a polarization image. After inferring two ambiguous normal maps, ($\mathbf{m}_1$, $\mathbf{m}_2$), as physical priors from the polarization image (see Sec.~\ref{sec:normal-estimation} for details), a two-branch strategy is adopted: one branch classifies each image pixel as belonging to either of the two normal maps or the background, thus obtaining the fused normal $\mathbf{m}_3$; a second branch regresses the coarse normal map $\mathbf{m}_4$ as an intermediate result. They are followed by the final step, which focuses on the residual refinement of coarse normal map to integrate the fused and the coarse normal maps as well as their cosine similarity, i.e. the normal residual, and to regress the final surface normal. Note that modules in the gray dash-line box are specifically designed for polarization images, to leverage the physical prior knowledge that reflected light from an object is polarized; these modules are unfit in dealing with RGB images.
    }
    \label{fig:polar2normal}
\end{figure*}

\subsection{Special Orthogonal Group} 
Mathematically, a Lie group~\cite{murray1994mathematical} is a group as well as a smooth manifold. 3D rotation transformations, also known as the Special Orthogonal group $\text{SO}(3)$, is exactly a Lie group and could be characterized by
\begin{equation}
    \text{SO}(3) = \{R\in\mathbb{R}^{3\times 3} | R^{\top}R=I, \mathrm{det}(R)=+1\}.
\end{equation}
The tangent space of Lie group $\text{SO}(3)$ at identity $\mathrm{I}_3$ is referred to as its Lie algebra $\mathfrak{so}(3)$. An element of $\mathfrak{so}(3)$ is a $3\times3$ skew-symmetric matrix $\hat W$, as
\begin{equation}
    \hat W=
    \begin{pmatrix}
        0 & -w_3 & w_2 \\
        w_3 & 0 & -w_1 \\
        -w_2 & w_1 & 0
    \end{pmatrix}.
\end{equation}
Essentially, $\mathfrak{so}(3)$ spans a 3-dimensional vector space, $\mathbf{w}=(w_1, w_2, w_3)^{\top}$. The mapping from a Lie algebra vector $\hat W\in \mathfrak{so}(3)$ to a point in the manifold $R\in\text{SO}(3)$ is formulated as an exponential map $\exp :\mathfrak{so}(3) \rightarrow \text{SO}(3)$ as
\begin{equation}
    R = \exp{(\hat W)}=\mathrm{I} + \frac{\sin(\|\mathbf{w}\|)}{\|\mathbf{w}\|}\hat W + \frac{1-\cos{(\|\mathbf{w}\|)}}{\|\mathbf{w}\|}\hat W^2,
\end{equation}
where $\|\cdot\|$ denotes the vector norm. The geodesic distance of two points in the manifold, $R_1, R_2 \in \text{SO}(3)$, is defined as the angular difference between the two rotations, which is
\begin{equation}
    D(R_1, R_2) = \Big|\cos^{-1}\Big( \frac{\Tr(R_1^{\top}R_2)-1}{2} \Big)\Big|.
\end{equation}

\section{Our HumanSfP Approach}
There are two main stages in our approach: (1) Polar2Normal: surface normal estimation from a single polarization image; (2) Polar2Shape: initial human body shape estimation from the raw polarization image and the obtained surface normal, which is followed by body shape refinement from the surface normal.

\subsection{Polar2Normal: Surface Normal Estimation}
\label{sec:normal-estimation}

The polarization image in our work consists of four channels, and each channel corresponds to one of the four polarizer angles: $0^{\circ}$, $45^{\circ}$, $90^{\circ}$ and $135^{\circ}$. Taking into account the $\pi$-ambiguity of $\varphi$, we have two possible solutions to the surface normal for each pixel, that form the physical priors, denoted as ambiguous normal maps $\mathbf{m}_1(\varphi, \theta)$ and $\mathbf{m}_2(\pi+\varphi, \theta)$. We propose a two-step architecture to estimate the surface normal from the polarization image. The details are presented in Fig.~\ref{fig:polar2normal}. 

In the first step, two encoders are used to extract visual features from the polarization image and the two ambiguous normal maps separately, which are then followed two decoders. One decoder is to capture a coarse surface normal of the human body denoted by $\mathbf{m}_3$, where Huber loss is employed to train the network, defined as
\begin{equation}
H(x, \alpha)=\left\{
\begin{aligned}
 & 0.5 \, x^2, & |x| < \alpha \\
 & \alpha \, (|x| - 0.5 \, \alpha), & \text{otherwise}
\end{aligned}
\right.
\end{equation}
The loss of the coarse surface normal becomes
\begin{equation}
    \label{eq:coarse-normal}
    \mathcal{L}_{\text{coarse}} = \sum_{i,j} H\big(1-\langle\mathbf{m}_{3 [i, j]}, \mathbf{m}_{[i,j]}\rangle, \alpha\big),
\end{equation}
where $\mathbf{m}$ is the target normal map with $(i, j)$ being the pixel coordinate, $\langle\mathbf{m}_{3 [i, j]}, \mathbf{m}_{[i,j]}\rangle$ is the cosine similarity between the two normal vectors, and $\alpha$ controls the trade-off between the squared and the absolute losses. 
Now, the second decoder is to classify each pixel into three categories: background, ambiguous normal $\mathbf{m}_1$, and ambiguous normal $\mathbf{m}_2$, with the corresponding pixel-wise probabilities $p_0$, $p_1$, and $p_2$, respectively. The fused normal is thus obtained by
\begin{equation}
    \label{eq:fuse-noisy-normal}
    \mathbf{m}_4 = (1-p_0) \cdot \frac{p_1 \mathbf{m}_1 + p_2 \mathbf{m}_2}{\|p_1 \mathbf{m}_1 + p_2 \mathbf{m}_2\|_2},
\end{equation}
where $1-p_0$ acts as a soft mask for the foreground human body. The classification loss is measured by the cross entropy between the predicted pixel-wise category and the target category,
\begin{equation}
    \label{eq:category-loss}
    \mathcal{L}_{\text{category}} = \sum_{i, j}\sum_c y_{c[i, j]}\log p_{c[i,j]} + (1-y_{c[i, j]})\log(1-p_{c[i,j]}).
\end{equation}
Here $c$ indexes among the three categories. $y_{c[i, j]}\in\{0,1\}$ is the multi-class label indicating which category the pixel $[i,j]$ belongs to. Note that the label $y_{c[i, j]}$ is created by discriminating whether the pixel is background or which ambiguous normal has higher cosine similarity with its target normal.

Now let us look at the second step. Different from our previous work~\cite{zou2020detailed} that directly regresses the normal map given the polarization image and fused normal, we propose a residual update scheme to produce a more detailed and accurate surface normal estimation, as displayed in Fig.~\ref{fig:polar2normal}. Due to the environmental noise and the digital quantization of the polarization image formation process, the fused normal map $\mathbf{m}_3$ is often noisy and non-smooth. Given the coarse normal map $\mathbf{m}_3$ and fused normal map $\mathbf{m}_4$, the normal residual $\mathbf{m}_{r}$ is evaluated by
\begin{equation}
    \mathbf{m}_{r[i,j]} = 1 - \langle\mathbf{m}_{3 [i, j]}, \mathbf{m}_{[i,j]}\rangle.
\end{equation}
A denoising network is then trained to take both normal maps $\mathbf{m}_3, \mathbf{m}_4$ and the normal residual $\mathbf{m}_{r}$ as input, to produce a smoothed and detailed normal $\mathbf{\hat m}$. 
The residual update scheme takes place as the coarse normal $\mathbf{m}_3$ is replaced by the final predicted normal $\mathbf{\hat m}$ to produce the new normal residual $\mathbf{m}_{r}$, which is then employed to regress an updated normal map $\mathbf{\hat m}$. The loss for $\mathbf{\hat m}$ is defined by
\begin{equation}
    \label{eq:normal-loss}
    \mathcal{L}_{\text{final}} = \sum_{i, j} \big|1-\langle\mathbf{\hat m}_{[i, j]}, \mathbf{m}_{[i,j]}\rangle\big|.
\end{equation}
Here $|\cdot|$ denotes the absolute value.
Finally, our surface normal estimation model is learned by minimizing the following loss
\begin{equation}
    \label{eq:polar2normal-loss}
    \mathcal{L}_{\text{polar2normal}} = \mathcal{L}_{\text{coarse}} + \mathcal{L}_{\text{category}} + \mathcal{L}_{\text{final}}.
\end{equation}

\subsection{Polar2Shape: Human Shape Reconstruction}

\begin{figure}
    \centering
    \includegraphics[width=\columnwidth]{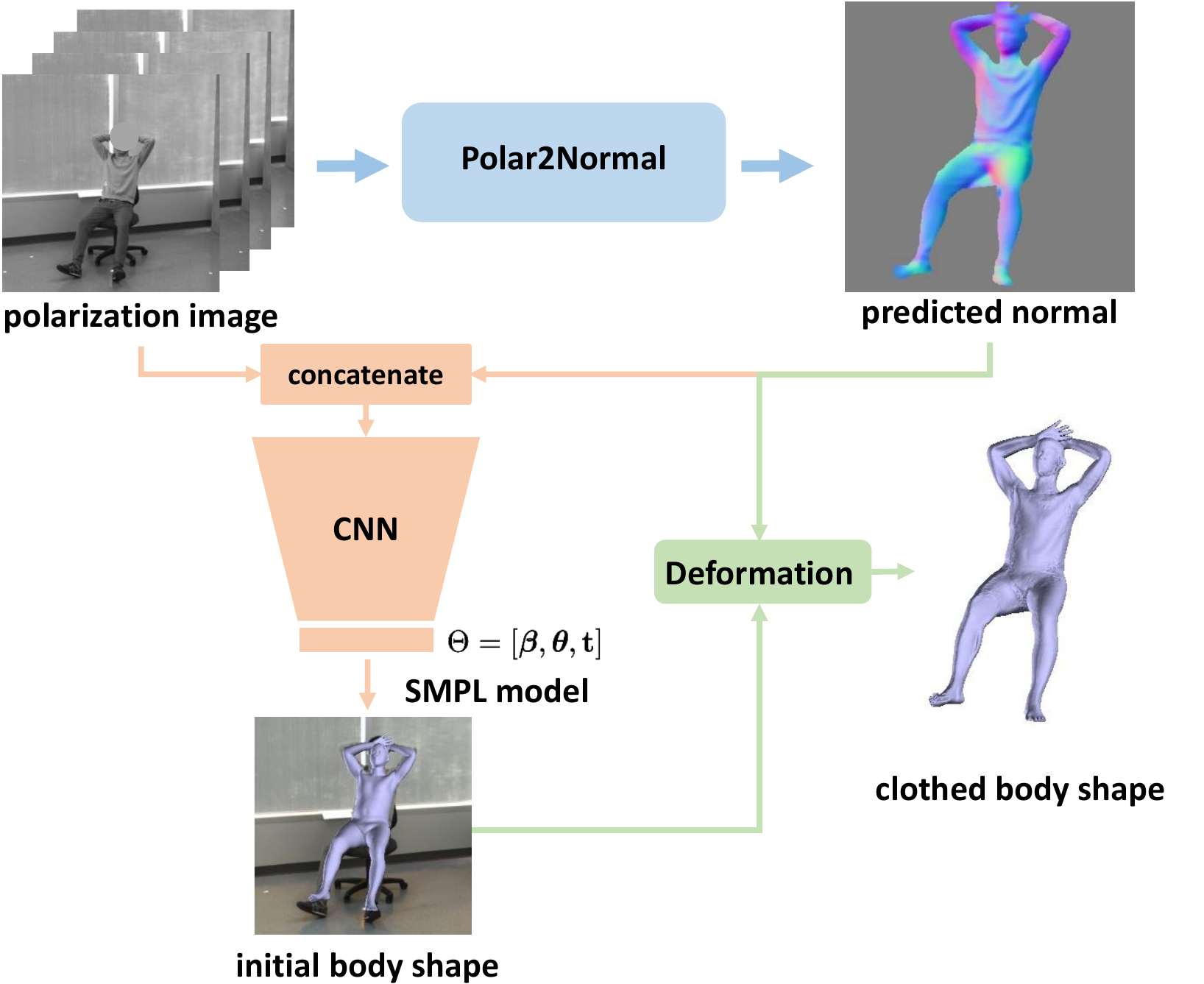}
    \caption{Stage two: our Polar2Shape pipeline of clothed body shape reconstruction from a polarization image, accomplished in two steps. The first step focuses on estimating the parameters of SMPL model, a rough \& naked shape model parameterized by $\Theta$. The next step is to deform the initial SMPL shape according to the estimated surface normal in Sec.\ref{sec:normal-estimation}, in reconstructing the refined 3D human shape with clothing details.}
    \label{fig:polar2shape}
\end{figure}

Stage two of our approach, also referred to as Polar2Shape, focuses on the reconstruction of clothed body shape from a polarization image, which consists of two steps. In the first step, the naked body shape, represented by SMPL model~\cite{loper2015smpl}, is estimated. The following step is to deform the initial human body shape with the estimated surface normal, and finally the clothed body shape is reconstructed.

\textbf{Initial Shape Estimation.} The core of SMPL model~\cite{loper2015smpl} lies in a differentiable function $\mathcal{M}(\boldsymbol{\beta}, \boldsymbol{\theta})\in\mathbb{R}^{6,890\times3}$ that outputs a triangular mesh with 6,890 vertices from 82 parameters $[\boldsymbol{\beta}, \boldsymbol{\theta}]$. $\boldsymbol{\theta}\in\mathbb{R}^{72}$ is the pose parameter vector to characterize pose articulations in axis-angle representation, consisting of one global rotation of the body and the relative rotations of its 23 joints. Human pose is therefore represented as $\boldsymbol{\theta}=(\boldsymbol{\theta}_m)_{m=1}^{24}$, where $\boldsymbol{\theta}_m\in\mathbb{R}^3$, and $m$ denotes the index of relative rotation in axis-angle. The shape parameter vector $\boldsymbol{\beta}\in\mathbb{R}^{10}$ is the linear coefficients of a PCA shape space that mainly determines individual body features such height, weight and body proportions. The PCA shape space is learned from a large dataset of naked and minimal clothed human body scans. A specific SMPL shape is produced by first applying pose-dependent and shape-dependent deformations to the template pose, then using forward-kinematics to articulate the body to its current pose, and deforming the surface mesh by linear blend skinning. At the same time, the 3D joint positions, denoted by $\mathbf{J}_{\text{3D}}\in\mathbb{R}^{24\times3}$, are obtained by linear regression from the output mesh vertices. In addition to the SMPL model parameters, the global translation of the human body is also a necessary factor in aligning with the projection in the 2D image space, denoted by $\mathbf{t}\in\mathbb{R}^3$. This results in an 85-dimensional parameter space, $\Theta=[\boldsymbol{\beta}, \boldsymbol{\theta}, \mathbf{t}]$, that are used by SMPL to dictate a specific human shape. 

Under mild assumption, the axis-angle representation of human pose in SMPL, $(\boldsymbol{\theta}_m)_{m=1}^{24}$, has a bijective map to the corresponding 24-dim product space of $\mathfrak{so}(3)$ manifold. Existing efforts such as~\cite{zou2020detailed} normally predict the pose in axis-angle representation, and measure the Euclidean distance between the predicted pose and target pose as $\sum_{m=1}^{24} \|\boldsymbol{\hat \theta}_m - \boldsymbol{\theta}_m\|_2$. However, as the two poses corresponds to two points in the aforementioned curved space, their distance is better characterized by their geodesic distance, which is not necessarily the Euclidean distance. Thus we propose to represent the human pose as a set of rotation matrices $\{R_m\}_{m=1}^{24}$ in $\text{SO}(3)$ with $R_m=\exp{(\boldsymbol{\theta}_m)}$. The geodesic distance between the predicted pose and target pose becomes
\begin{equation}
    \mathcal{L}_{\text{pose}} = \sum_{m=1}^{24} D(R_m, \hat R_m) = \sum_{m=1}^{24} \Big| \cos^{-1} \Big( \frac{\Tr(R_m^{\top}\hat R_m)-1}{2} \Big)\Big|,
\end{equation}
which is also referred as the pose estimation loss.
Moreover, the losses for SMPL shape, global translation and joint positions are defined as
\begin{align}
    \mathcal{L}_{\text{shape}} &= \|\boldsymbol{\beta} - \boldsymbol{\hat\beta}\|_2^2,\\
    \mathcal{L}_{\text{trans}} &= \|\mathbf{t} - \mathbf{\hat t}\|_2^2,\\
    \mathcal{L}_{\text{joints}} &= \|\mathbf{J}_{\text{3D}} - \mathbf{\hat J}_{\text{3D}}\|_2^2.
\end{align}
At the moment, our initial shape estimation is learned by minimizing the following loss
\begin{equation}
    \mathcal{L}_{\text{polar2shape}} = \mathcal{L}_{\text{pose}} + \lambda_{\text{shape}}\mathcal{L}_{\text{shape}} + \lambda_{\text{trans}}\mathcal{L}_{\text{trans}} + \lambda_{\text{joints}}\mathcal{L}_{\text{joints}},
\end{equation}
where $\lambda_{\text{shape}}$, $\lambda_{\text{trans}}$ and $\lambda_{\text{joints}}$ are the tuning parameters of the corresponding loss terms.

\textbf{Shape Reconstruction.} The initial human shape obtained by SMPL representation still lacks fine surface details. Therefore, the aim of this step is to refine the initial SMPL shape guided by our surface normal estimate, as follows. The SMPL body shape is rendered on the image plane to form an initial depth map. The technique of~\cite{nehab2005efficiently} is then engaged here to obtain an optimized depth map $I_d$ from the predicted surface normal $\mathbf{\hat m}$ and the initial depth $\hat I_d$ map by minimizing the objective function,
\begin{equation}
    E(I_d) = \lambda_n E_n(I_d) + \lambda_d E_d(I_d) + \lambda_s E_s(I_d),
    \label{eq:integrate}
\end{equation}
which contains three energy terms. The first term, $E_n(I_d)$, ensures the predicted normal to be perpendicular to the tangents of the optimized depth surface. The second term, $E_d(I_d)$, encourages the optimized depth to be close to the initial depth. The third and final term preserves smoothness of nearby pixels over the optimized depth map. More details can be found in the supplementary material.

\subsection{Our In-house PHSPD Dataset}
To facilitate empirical evaluation of our approach in real-world scenarios, a home-grown dataset is curated, which is referred as Polarization Human Pose and Shape Dataset or PHSPD. In what follows, we start by presenting the early version, PHSPDv1, as well as the more recent addition, PHSPDv2.

\textbf{PHSPDv1.} It is the early version used in our preliminary work~\cite{zou2020detailed}. During PHSPDv1 data acquisition, a system of 4 soft-synchronized cameras are used, consisting of a polarization camera and three RGBD cameras. 12 subjects are recruited in data collection, in which 9 are male and 3 are female. Each subject performs 3 different sets of actions (out of 18 distinct action types) 4 times, plus an additional period of free-form motion at the end of each session. Thus for each subject, there are 13 short videos (around 1,800 frames per video with 10-15 FPS). The total number of frames for each subject amounts to 22K. Overall, PHSPDv1 dataset consists of 287K frames, each frame here contains a synchronized set of images: one polarization image, three RGBD images.

\textbf{PHSPDv2.} It contains the newly acquired data, where our multi-camera acquisition system is extended to 11 cameras of different modality: one polarization camera and five RGBD cameras. 15 subjects are recruited for the data acquisition, where 11 are male and 4 are female. Each subject is required to perform 3 groups of actions (21 different actions in total) for 4 times, where each group includes actions of fast/medium/slow speed, respectively. Finally, 12 videos are collected for each subject; each video has around 1,300 frames of 15 FPS. In total, there are 180 videos, with each video lasting about 1.5 minutes. This amounts to 240k frames with each frame including a polarization image and five RGBD images. 

To summarize, our PHSPD dataset contains 334 videos of 21 different subjects performing 31 types of actions, amounts to around 527K frames and 9.5 hours of recorded videos. Each frame contains a synchronized set of images including both a polarization image and RGBD images. Further details regarding our PHSPD dataset can be found in the supplementary material.

\textbf{Annotation.} The human full-body 3D joint positions and SMPL skinning weight parameters are obtained from the multiple RGBD cameras as follows. For each frame, the 2-D joints of all the RGB images are detected by OpenPose~\cite{openpose}; the coarse depth of each 2-D joint is obtained by RGBD camera SDK. After aggregating the multi-view initial 3-D joint positions, the SMPLify-x~\cite{pavlakos2019smplifyx} model is fitted to the initial pose. For more precise annotations, the initial shape is then fine-tuned to fit the point cloud collected from multi-view depth images using the L-BFGS~\cite{bollapragada2018LBFGS} optimizer, where the average distance of shape vertex to its nearest point in the point cloud is iteratively minimized. Exemplar annotated human shapes can be found in the supplementary material.

\begin{table}[htp]
    \centering
    \begin{tabular}{|p{58pt}|p{13pt}p{12pt}p{15pt}p{18pt}p{22pt}p{10pt}p{7pt}|}
    \bottomrule \hline
        \makecell[c]{Dataset} & \makecell[c]{Sub} & \makecell[c]{Act} & \makecell[c]{MM} & \makecell[c]{RGB} & \makecell[c]{Depth} & \makecell[c]{P} & \makecell[c]{S}\\
    \hline
        \makecell[c]{MS COCO~\cite{lin2014microsoft}} & \makecell[c]{-} & \makecell[c]{-} & \makecell[c]{\xmark} & \makecell[c]{330K} & \makecell[c]{\xmark} & \makecell[c]{2D} & \makecell[c]{\xmark} \\
        \makecell[c]{MPII~\cite{akhter2015pose}} & \makecell[c]{-} & \makecell[c]{-} & \makecell[c]{\xmark} & \makecell[c]{40K} & \makecell[c]{\xmark} & \makecell[c]{2D} & \makecell[c]{\xmark} \\
        \makecell[c]{PoseTrack~\cite{andriluka2018posetrack}} & \makecell[c]{-} & \makecell[c]{-} & \makecell[c]{\xmark} & \makecell[c]{22K} & \makecell[c]{\xmark} & \makecell[c]{2D} & \makecell[c]{\xmark} \\
        \makecell[c]{MPI-3DHP~\cite{mehta2017monocular}} & \makecell[c]{8} & \makecell[c]{-} & \makecell[c]{\xmark} & \makecell[c]{1.3M} & \makecell[c]{\xmark} & \makecell[c]{3D} & \makecell[c]{\xmark} \\
        \makecell[c]{3DPW~\cite{von2018recovering}} & \makecell[c]{7} & \makecell[c]{8} & \makecell[c]{\xmark} & \makecell[c]{51K} & \makecell[c]{\xmark} & \makecell[c]{3D} & \makecell[c]{\cmark} \\
        \makecell[c]{Human3.6M~\cite{human36m}} & \makecell[c]{11} & \makecell[c]{15} & \makecell[c]{\xmark} & \makecell[c]{3.6M} & \makecell[c]{0.45M} & \makecell[c]{3D} & \makecell[c]{\xmark} \\
        \makecell[c]{PHSPD~(ours)} & \makecell[c]{21} & \makecell[c]{31} & \makecell[c]{\cmark} & \makecell[c]{2.1M} & \makecell[c]{2.1M} & \makecell[c]{3D} & \makecell[c]{\cmark} \\
    \hline \toprule 
    \end{tabular}
    \caption{A tally of widely-used human pose and shape datasets, compared in terms of number of subjects (Sub), number of different actions (Act), multi-modality (MM), number of RGB (RGB) and depth (Depth) frames, availability of annotated poses (P) and shapes (S) for each frame.}
    \label{tab:dataset-compare}
\end{table}

\textbf{Comparison with Existing Datasets.} Our PHSPD dataset is compared side-by-side with six widely-used human pose and shape datasets in Table~\ref{tab:dataset-compare}. The datasets of MPII~\cite{akhter2015pose}, MS COCO~\cite{lin2014microsoft} and PoseTrack~\cite{andriluka2018posetrack} provide only RGB images with 2D pose annotations, where depth images and shape annotations are not considered; MPI-INF-3DHP~\cite{mehta2017monocular} contains 1.3M in-house frames with 3D pose annotations, but without depth channel images. 3DPW~\cite{von2018recovering} has 51K frames with extreme pose and 3D shape annotations, meanwhile it does not possess depth images, it is also has a limited amount of annotated images. Comparing to Human3.6M, our PHSPD dataset comes with 3D shape annotations, RGB images of higher resolution, more depth images and with higher resolution, more subjects, and more action types. More importantly, PHSPD is the only dataset that comes with polarization images.

\section{Experiments}
Our proposed approach is empirically examined in two major aspects, namely surface normal estimation, and pose and shape estimation from a polarization image. Sec.~\ref{sec:exp-normal-estimation} focuses on the empirical evaluations of surface normal estimation on the widely used SfP benchmark~\cite{ba2019physics} in shape from polarization, as well as our PHSPD dataset. Sec.~\ref{sec:exp-pose-estimation} and~\ref{sec:exp-shape-estimation} evaluate 3D pose estimation on the synthetic SURREAL dataset~\cite{varol2017learning} and the PHSPD dataset, and shape estimation on our PHSPD dataset. Ablation study is presented in Sec.~\ref{sec:ablation} to analyze the effect of individual components in our approach.

\textbf{Evaluation Metrics.} For surface normal estimation, we report mean angle error (MAE), which measures the angle error between the target and estimated normal map. For human pose and shape estimation, we report the mean per joint position error (MPJPE), pelvis-aligned MPJPE (PEL-MPJPE) where two pelvis joints are aligned between the predicted and target 3D pose, Procrustes-aligned MPJPE (PA-MPJPE) where the predicted pose is aligned with the target by rigid transformation, and percentage of correct key-points (PCK) where the joint position error less than 100mm is considered as the correct prediction. As for the evaluation of human shape, 3D point to surface error (P2S) is employed, where the iterative closest point (ICP) alignment between predicted body mesh and the ground-truth human body point cloud is applied; for each vertex of the human body triangle mesh, its closest point in the point cloud is identified to form a pair, and the average distance of the pairs is computed.

\textbf{Evaluation Datasets.} 
The widely-used SfP dataset~\cite{ba2019physics} is employed to evaluate the performance of our proposed Polar2Normal component. We follow the typical training (236 polarization images with 1224x1024 resolution) and testing (27 images) scheme as in~\cite{ba2019physics}, where a 256x256 patch on a image is randomly cropped for training, and 20 overlapped patches of one image are first cropped for testing and then fused to form the final predicted normal map. 
We also demonstrate the effectiveness of our approach on SURREAL~\cite{varol2017learning}, a synthetic dataset of color images rendered from motion-captured SMPL human shapes. Polarization images can be synthesized using color and depth images (details are in supplementary material). We choose subset "run1" and subsample over time to recruit one from every 20 consecutive frames. Finally, the train set comprises 123,860 samples and test set has 26,650 samples. SURREAL dataset is only used for the evaluation of normal and pose estimation due to the lack of ground truth point cloud of the human body. 
Our real-world PHSPD dataset is involved in the evaluations of all three tasks of normal estimation, pose estimation and clothed shape estimation. The target surface normal is obtained by fusing multi-view normal maps calculated from multiple depth images. Subject 4, 7, 11 in PHSPDv1 and subject 1, 2, 7 in PHSPDv2 are chosen to form the test set, resulting in 117,860 samples. The training set consists of the rest 400,785 samples.

\textbf{Implementation Details.} For Polar2Normal, each of the encoders and decoders contains 6 sequential blocks, with each block having one up/down-sampling and two convolutional layers. Further details of our Polar2Normal architecture can be found in supplementary material. The Polar2Normal model is trained for 600 epochs on SfP dataset, 30 epochs on SURREAL dataset and 8 epochs on PHSPD dataset with Adam optimizer~\cite{kingma2014adam}. The learning rate is 0.001, which decays to 0.0001 after training 200 epochs, 10 epochs, and 6 epochs, respectively. The batch size is 16 for training on three datasets. For Polar2Shape model, ResNet50~\cite{he2016deep} is used as the backbone CNN model. The extract 1024-dimensional feature is directly regressed to the final outputs: $\boldsymbol{\beta}$, $\boldsymbol{\theta}$, and $\mathbf{t}$. The pose estimation model is trained for 30 epochs on SURREAL dataset and 8 epochs on PHSPD dataset with Adam optimizer~\cite{kingma2014adam}. The learning rate is 0.001 and decays to 0.0001 after training 10 epochs and 4 epochs respectively. The batch size for training is 32. The trade-off parameter $\alpha$ is set to 0.5. 
The tuning parameters of the loss terms are set to $\lambda_{\text{shape}}=0.1$, $\lambda_{\text{trans}}=0.1$ and $\lambda_{\text{joints}}=10$, respectively. The three weights used in the Polar2Shape stage, namely the normal term $\lambda_n$, the depth data term $\lambda_d$, and the smoothness term $\lambda_s$ are empirically set to 1.0, 0.06, and 0.55, respectively.

\subsection{Evaluation of Surface Normal Estimation}
\label{sec:exp-normal-estimation}
\begin{figure}
    \centering
    \includegraphics[width=\columnwidth]{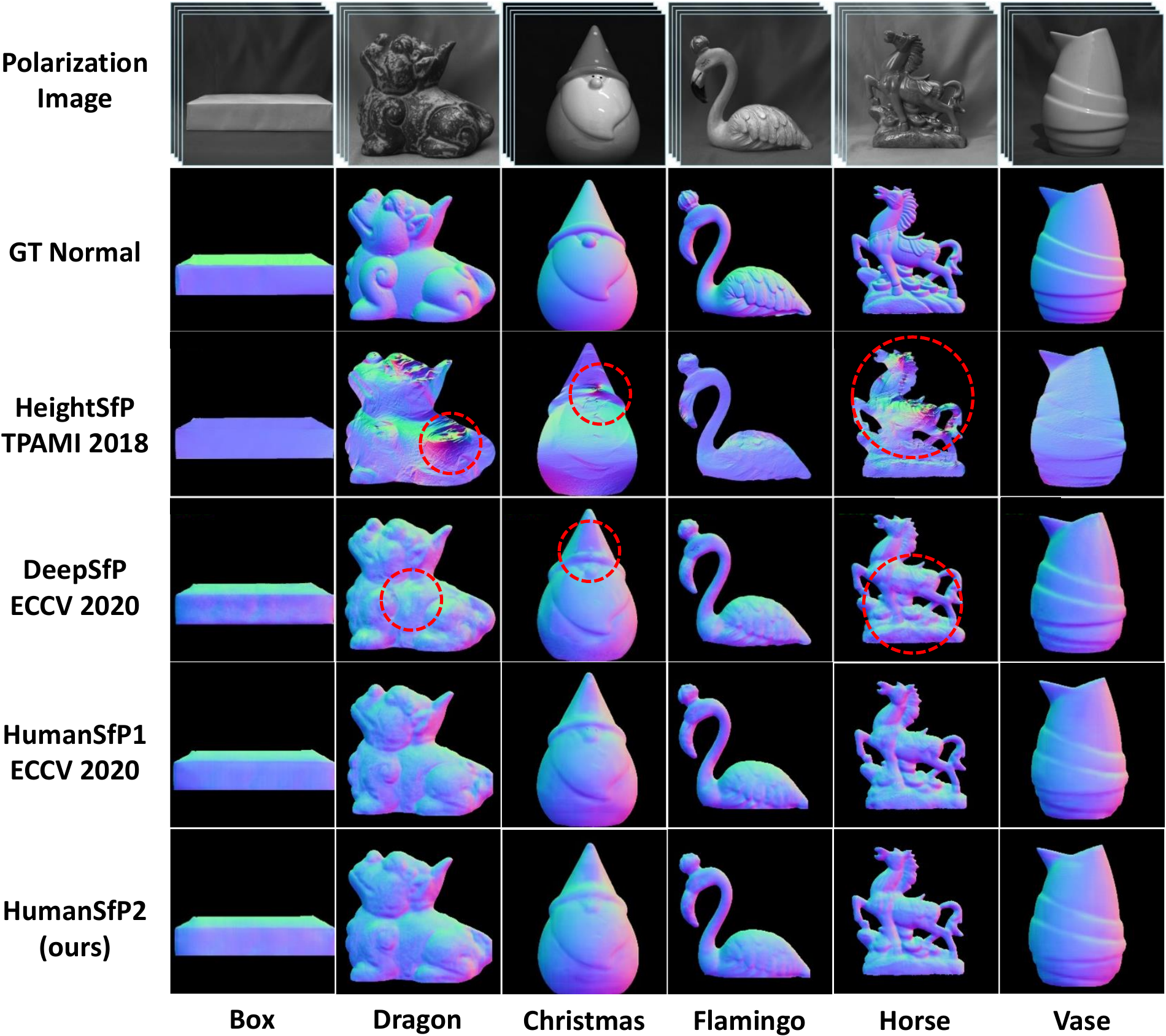}
    \caption{Exemplar results of normal map prediction on SfP dataset by four comparison methods: HeightSfP~\cite{smith2016linear}, DeepSfP~\cite{ba2019physics}, HumanSfP1~\cite{zou2020detailed}, and HumanSfP2~(ours). Original polarization images and GT normal maps are shown in the first and second row.}
    \label{fig:sfp-normal-estimation-demo}
\end{figure}

\begin{figure*}
    \centering
    \includegraphics[width=0.75\textwidth]{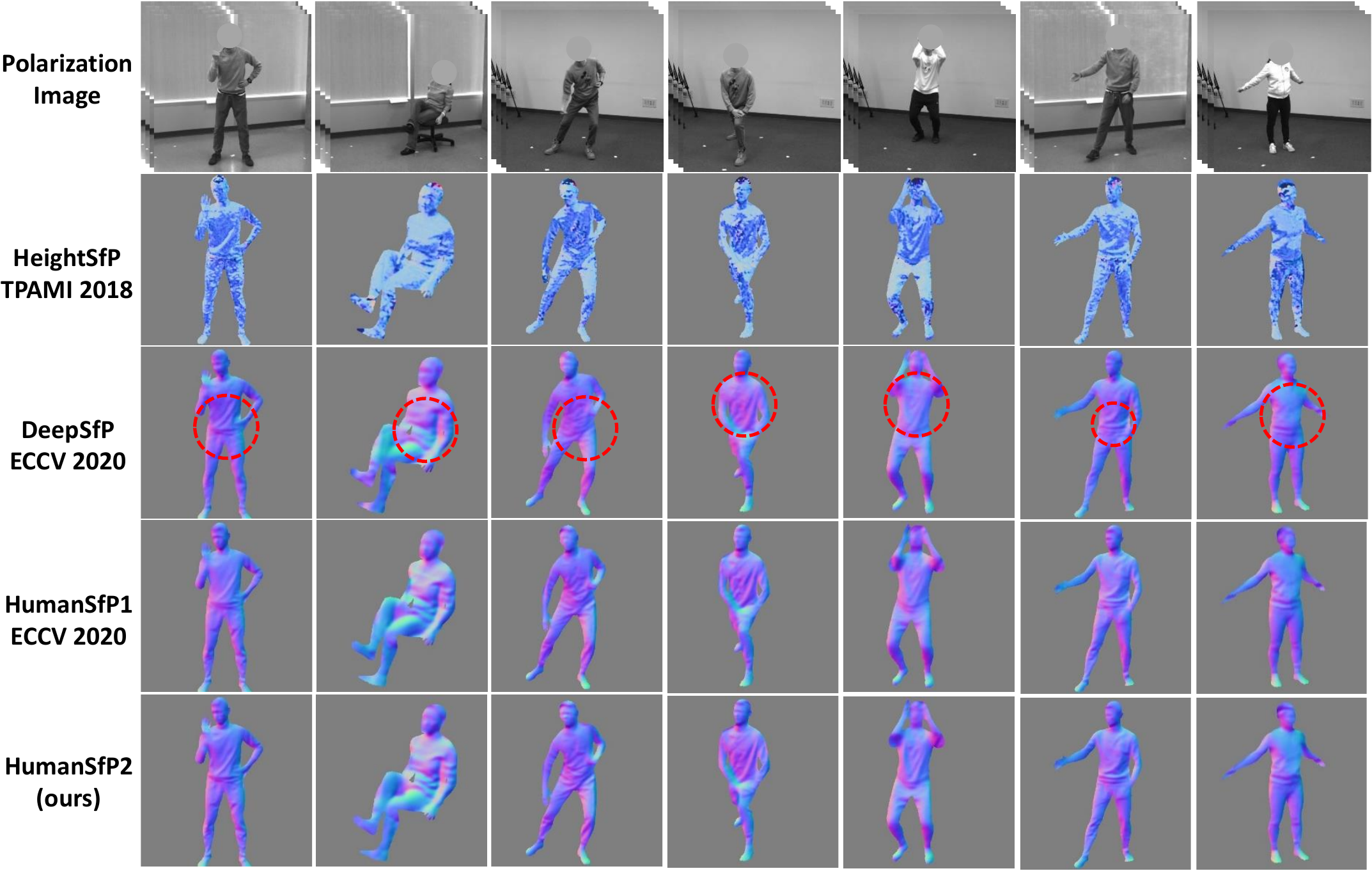}
    \caption{Exemplar results of normal map prediction on PHSPD dataset from four comparison methods: HeightSfP~\cite{smith2016linear}, DeepSfP~\cite{ba2019physics}, HumanSfP1~\cite{zou2020detailed}, and HumanSfP2~(ours).}
    \label{fig:normal-estimation-demo}
\end{figure*}

In this task, our approach is compared with three baselines: a conventional method HeightfP~\cite{smith2016linear}, a most recent work DeepSfP~\cite{ba2019physics} and our preceding work HumanSfP1~\cite{zou2020detailed}. 

From the quantitative results in Table~\ref{tab:normal-estimation-sfp} and the qualitative results in Fig.~\ref{fig:sfp-normal-estimation-demo}, our method consistently outperforms the state-of-the-art SfP methods HeightfP~\cite{smith2016linear} and DeepSfP~\cite{ba2019physics} for the task of shape from polarization. The poor performance of HeightfP~\cite{smith2016linear} could be attributed to its noise-free assumption that may not hold in the captured images. Though DeepSfP~\cite{ba2019physics} incorporates the ambiguous normal maps as physical priors, the ambiguous normal maps are directly concatenated with the polarization image to form its input, which may overlook the implicit geometric clues. As a result, it performs less well when comparing to our method, especially in complex scenes such as Christmas, Dragon and Horse, where the results of our method achieve $\sim 3^{\circ}-5^{\circ}$ improvement. It is also demonstrated in the visual results of Fig.~\ref{fig:sfp-normal-estimation-demo}. Comparing to our previous work HumanSfP1~\cite{zou2020detailed}, as illustrated in Table~\ref{tab:normal-estimation-sfp}, our new method, HumanSfP2, exceeds HumanSfP1 in most scenarios, especially Horse and Dragon, where we have $\sim 2^{\circ}$ improvement; Only for the two scenes of Box and Christmas, the results of both HumanSfP1 and HumanSfP2 are almost identical. The observation demonstrates the advantages of our newly proposed 2-step strategy for surface normal estimation.

\begin{table}
    \centering
    \begin{tabular}{|c|p{38pt}p{38pt}p{40pt}p{40pt}|}
    \bottomrule \hline
        Scene & \makecell[c]{HeightfP \\ \cite{smith2016linear}} & \makecell[c]{DeepSfP \\ \cite{ba2019physics}} & \makecell[c]{HumanSfP1 \\ \cite{zou2020detailed}} & \makecell[c]{HumanSfP2 \\ (ours)}\\
    \hline
        Dragon & \makecell[c]{49.16} & \makecell[c]{21.55} & \makecell[c]{18.71} & \makecell[c]{\textbf{16.88}} \\
        Horse & \makecell[c]{55.87} & \makecell[c]{22.27} & \makecell[c]{21.27} & \makecell[c]{\textbf{19.64}} \\
        Christmas & \makecell[c]{39.68} & \makecell[c]{13.50} & \makecell[c]{\textbf{8.56}} & \makecell[c]{8.57} \\
        Box & \makecell[c]{31.00} & \makecell[c]{23.31} & \makecell[c]{\textbf{21.94}} & \makecell[c]{22.04} \\
        Flamingo & \makecell[c]{36.05} & \makecell[c]{\textbf{20.19}} & \makecell[c]{20.51} & \makecell[c]{20.44} \\
        Vase & \makecell[c]{36.88} & \makecell[c]{10.32} & \makecell[c]{9.20} & \makecell[c]{\textbf{9.18}} \\
        Whole Set & \makecell[c]{41.44} & \makecell[c]{18.52} & \makecell[c]{17.22} & \makecell[c]{\textbf{16.73}} \\
    \hline \toprule 
    \end{tabular}
    \caption{Quantitative evaluations of surface normal estimation on SfP dataset~\cite{ba2019physics}. MAE is reported in the case of six scenes and the whole set. HumanSfP2~(ours) outperforms two baseline methods~\cite{smith2016linear,ba2019physics} by a large margin. It also demonstrates that our new method, HumanSfP2~(ours) outperforms the previous effort, HumanSfP1~\cite{zou2020detailed}, especially in the cases of complex shapes such as Horse and Dragon.}
    \label{tab:normal-estimation-sfp}
\end{table}

We also evaluate the normal estimation of human body surface on both SURREAL dataset and our PHSPD dataset. Through both the quantitative results of Table~\ref{tab:normal-estimation} and the visual results of Fig.~\ref{fig:normal-estimation-demo}, it is observed that our method consistently outperforms the state-of-the-art surface normal prediction methods, namely HeightfP~\cite{smith2016linear}, DeepSfP~\cite{ba2019physics} and HumanSfP1~\cite{zou2020detailed}. Similar to the results presented in Table~\ref{tab:normal-estimation-sfp}, the conventional method HeightfP shows least appealing performance, which possibly results from its assumptions of polarization images with high-precision pixel intensity and ideal noise-free environment. This is in contrast to the deep neural network based approaches that are often more robust to environmental noise and polarization images with standard pixel representation. Rather than directly concatenating and feeding the input polarization image and the corresponding ambiguous normal maps into a deep model, HumanSfP1 and HumanSfP2 categorize each pixel into one of the ambiguous normal maps, and obtain a fused normal that incorporates explicit geometric information for normal estimation. This could be the main factor that our methods exceed the state-of-the-art DeepSfP on both SURREAL and PHSPD datasets. Moreover, HumanSfP2 out-performs HumanSfP1 over both datasets, which may be attributed to the 2-step model proposed in HumanSfP2, where the first step focuses more on the coarse normal map, and the second step pays more attention to the fine normal details. Qualitative results in Fig.~\ref{fig:normal-estimation-demo} also demonstrate the superior results of the proposed HumanSfP2.

\begin{table}
    \centering
    \begin{tabular}{|c|p{50pt}p{45pt}|}
    \bottomrule \hline
        Method & \makecell[c]{SURREAL} & \makecell[c]{PHSPD} \\
    \hline
        HeightfP~\cite{smith2016linear} & \makecell[c]{20.03} & \makecell[c]{39.95} \\
        DeepSfP~\cite{ba2019physics} & \makecell[c]{7.59} & \makecell[c]{23.08} \\
        HumanSfP1~\cite{zou2020detailed} & \makecell[c]{7.08} & \makecell[c]{22.06} \\
    \hline
        HumanSfP2~(ours) & \makecell[c]{\textbf{6.79}} & \makecell[c]{\textbf{21.36}} \\
    \hline \toprule 
    \end{tabular}
    \caption{Quantitative evaluation of surface normal estimation on SURREAL and PHSPD datasets in terms of MAE. HumanSfP2~(ours) significantly outperforms the three baseline methods.}
    \label{tab:normal-estimation}
\end{table}

\subsection{Evaluation of Pose Estimation}
\label{sec:exp-pose-estimation}

\begin{figure*}
    \centering
    \includegraphics[width=0.95\textwidth]{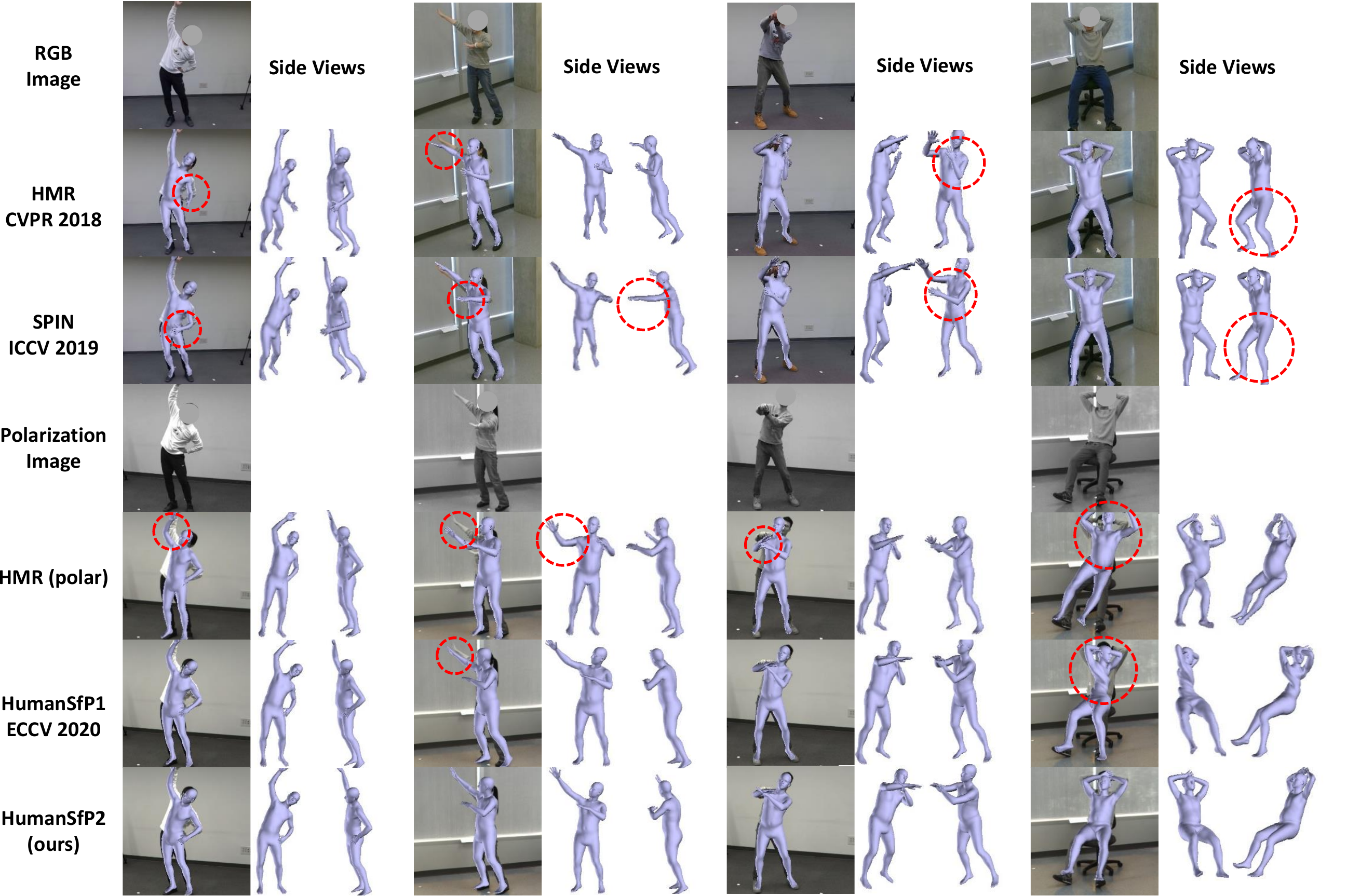}
    \caption{Exemplar results of human pose estimation on PHSPD dataset by five competing methods: HMR~\cite{kanazawa2018end}, SPIN~\cite{kolotouros2019learning}, HMR~(polar), HumanSfP1~\cite{zou2020detailed}, and HumanSfP2~(ours). The input RGB and polarization images are shown in the 1st and 4th row. Poses from two side views are also presented to better evaluate the predicted results.}
    \label{fig:pose-estimation-demo}
\end{figure*}

\begin{table*}
    \centering
    \begin{tabular}{|c|c|p{50pt}p{45pt}p{40pt}|p{40pt}p{50pt}p{45pt}p{40pt}|}
    \bottomrule \hline
        \multirow{2}{*}{Method} & \multirow{2}{*}{Input} & \multicolumn{3}{c|}{SURREAL}  & \multicolumn{4}{c|}{PHSPD} \\
        \cline{3-9}
        & & \makecell[c]{PEL-MPJPE $\downarrow$} & \makecell[c]{PA-MPJPE $\downarrow$} & \makecell[c]{PCK $\uparrow$} & \makecell[c]{MPJPE $\downarrow$} & \makecell[c]{PEL-MPJPE $\downarrow$} & \makecell[c]{PA-MPJPE $\downarrow$} & \makecell[c]{PCK $\uparrow$} \\
    \hline
        \makecell[c]{HMR~\cite{kanazawa2018end}} & \makecell[c]{C} & \makecell[c]{135.95} & \makecell[c]{100.66} & \makecell[c]{0.53} & \makecell[c]{-} & \makecell[c]{106.27} & \makecell[c]{68.58} & \makecell[c]{0.57} \\
        \makecell[c]{SPIN~\cite{kolotouros2019learning}} & \makecell[c]{C} & \makecell[c]{95.08} & \makecell[c]{80.67} & \makecell[c]{0.58} & \makecell[c]{-} & \makecell[c]{91.92} & \makecell[c]{49.79} & \makecell[c]{0.64} \\
        \makecell[c]{HMR~(polar)} & \makecell[c]{P} & \makecell[c]{113.82} & \makecell[c]{86.85} & \makecell[c]{0.55} & \makecell[c]{113.74} & \makecell[c]{88.24} & \makecell[c]{59.05} & \makecell[c]{0.68} \\
        \makecell[c]{HumanSfP1~\cite{zou2020detailed}} & \makecell[c]{P} & \makecell[c]{84.78} & \makecell[c]{60.82} & \makecell[c]{0.72} & \makecell[c]{66.97} & \makecell[c]{63.12} & \makecell[c]{42.08} & \makecell[c]{0.83} \\
        \makecell[c]{HumanSfP2~(ours)} & \makecell[c]{P} & \makecell[c]{\textbf{59.17}} & \makecell[c]{\textbf{46.58}} & \makecell[c]{\textbf{0.85}} & \makecell[c]{\textbf{62.04}} & \makecell[c]{\textbf{54.98}} & \makecell[c]{\textbf{36.88}} & \makecell[c]{\textbf{0.88}} \\
    \hline \toprule 
    \end{tabular}
    \caption{Quantitative evaluations of human pose estimation on both SURREAL and PHSPD datasets. The unit of joint error is in millimeter. HMR~\cite{kanazawa2018end} and SPIN~\cite{kolotouros2019learning} take RGB images (C) as the input, while HMR~(polar), HumanSfP1~\cite{zou2020detailed} and HumanSfP2~(ours) take polarization images (P) as the input.}
    \label{tab:pose-estimation-polarization}
\end{table*}

This section concerns the qualitative and quantitative evaluation of the estimated SMPL poses. Our comparison methods consist of HMR~\cite{kanazawa2018end} and SPIN~\cite{kolotouros2019learning}. Since HMR and SPIN is trained on single RGB images, for PHSPD dataset, images from an RGB camera having similar angle of view with the polarization camera are used as the input for evaluation. In addition to HMR and SPIN, for fair comparison, HMR~(polar) is included as another baseline, where HMR model is trained from scratch on the polarization images of either SURREAL dataset or our PHSPD dataset. HumanSfP1~\cite{zou2020detailed} is also employed as another baseline that uses Euclidean distance to measure the distance between the predicted and target poses.

From Table~\ref{tab:pose-estimation-polarization}, it is observed that our HumanSfP2 method produces the lowest errors in MPJPE, PEL-MPJPE and PA-MPJPE, and the highest PCK score among all competing methods. Comparing to HMR and SPIN that takes RGB images as the input source of data, and HMR~(polar) that takes polarization images as the input source, our HumanSfP2 out-performs them consistently by a large margin on both SURREAL and PHSPD datasets. Such superior performance lies in two important factors. The first is the engagement of normal map as part of the input, which will be explained in Sec.~\ref{sec:ablation} in detail. It is of interest to point out that normal map as part of the input data source is capable of reducing the average joint error by about $10-20$ mm in MPJPE, PEL-MPJPE and PA-MPJPE, while the three competing methods do not include the normal clues in pose estimation. The second factor is the introduction of geodesic loss to measure the predicted and target SMPL poses. Comparing to HumanSfP1, our approach improves the joint error by about $5$mm and PCK by $0.05$ in PHSPD dataset, and $15$mm and $0.13$ in SURREAL dataset. The quantitative results illustrate the effectiveness of leveraging geodesic distance in pose estimation. Visual results in Fig.~\ref{fig:pose-estimation-demo} also demonstrate the effectiveness of our HumanSfP2 approach, where HumanSfP1 may predict unnatural poses as the Euclidean distance is not suitable in measuring joint displacements due to relative joint rotations.

\subsection{Evaluation of Shape Estimation}
\label{sec:exp-shape-estimation}

\begin{figure*}
    \centering
    \includegraphics[width=\textwidth]{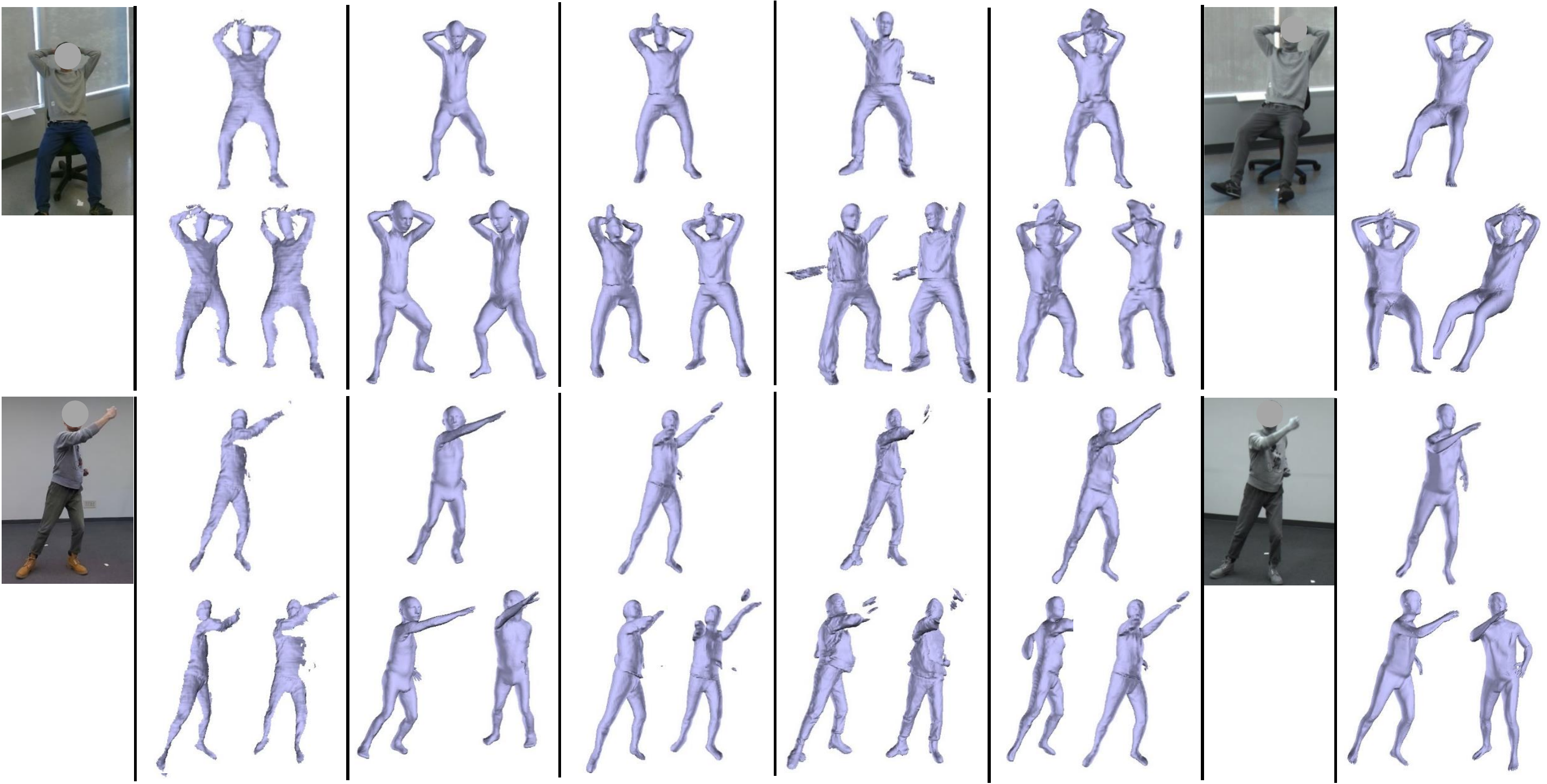}
    \includegraphics[width=\textwidth]{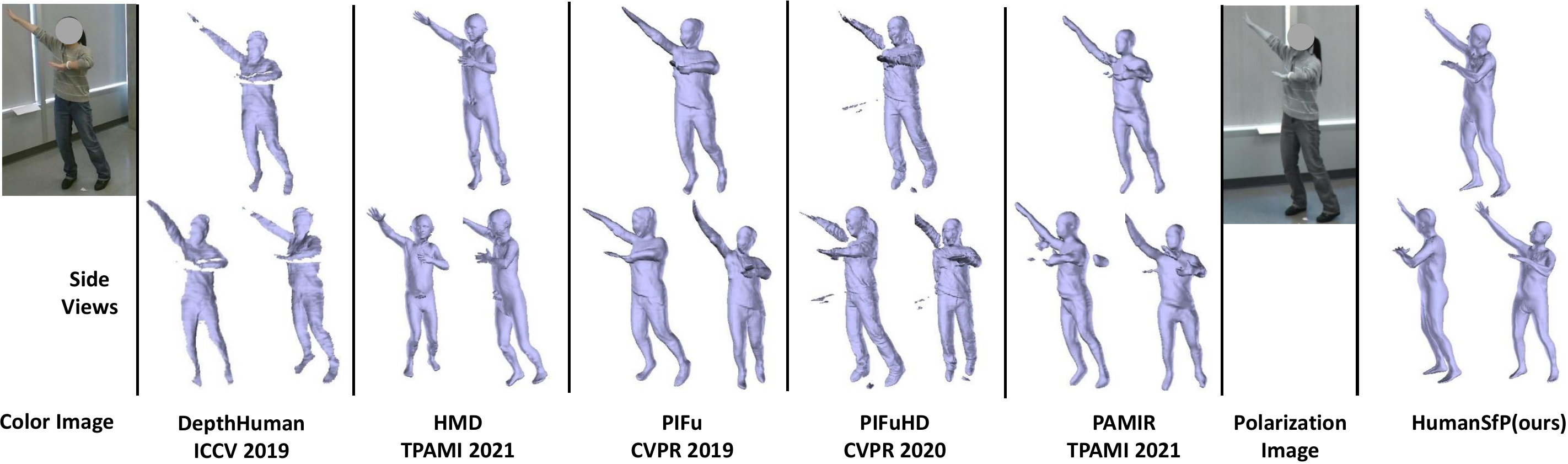}
    \caption{Exemplar estimation results of clothed body shapes. The first and sixth column are RGB and polarization images, respectively. DepthHuman~\cite{tang2019neural}, HMD~\cite{zhu2021detailed}, PIFu~\cite{saito2019pifu}, PIFuHD~\cite{saito2020pifuhd} and PaMIR~\cite{zheng2021pamir} produce results based on RGB images. HumanSfP~(ours) works with polarization input image.}
    \label{fig:detailed-shape-demo}
\end{figure*}

\begin{table}
    \centering
    \begin{tabular}{|c|c|p{50pt}|}
    \bottomrule \hline
        Method & Input & \makecell[c]{P2S}\\
    \hline
        DepthHuman~\cite{tang2019neural} & C & \makecell[c]{83.86}  \\
        PIFuHD~\cite{saito2020pifuhd} & C & \makecell[c]{67.05}  \\
        PIFu~\cite{saito2019pifu} & C & \makecell[c]{62.13}  \\
        HMD~\cite{zhu2021detailed} & C & \makecell[c]{43.72}  \\
        PaMIR~\cite{zheng2021pamir} & C & \makecell[c]{47.24}  \\
    \hline
        HumanSfP2~(initial) & P & \makecell[c]{39.79} \\
        HumanSfP2~(ours) & P & \makecell[c]{\textbf{38.24}} \\
    \hline \toprule 
    \end{tabular}
    \caption{Quantitative evaluations of clothed human shape estimation. DepthHuman~\cite{tang2019neural}, PIFuHD~\cite{saito2020pifuhd}, PIFu~\cite{saito2019pifu}, HMD~\cite{zhu2021detailed} and PaMIR~\cite{zheng2021pamir} take RGB images as input, while HumanSfP2~(initial) and HumanSfP2~(ours) take as input polarization images.}
    \label{tab:shape-estimation}
\end{table}

\begin{table}
    \centering
    \begin{tabular}{|c|p{50pt}p{45pt}|}
    \bottomrule \hline
        Method & \makecell[c]{SURREAL} & \makecell[c]{PHSPD} \\
    \hline
        HumanSfP2~(color) & \makecell[c]{13.44} & \makecell[c]{24.79} \\
        HumanSfP2~(no-prior) & \makecell[c]{12.16} & \makecell[c]{24.75} \\
    \hline
        HumanSfP2~(ours) & \makecell[c]{\textbf{6.79}} & \makecell[c]{\textbf{21.36}} \\
    \hline \toprule 
    \end{tabular}
    \caption{Ablation study of our normal estimation component on SURREAL and PHSPD datasets.}
    \label{tab:normal-estimation-ablation}
\end{table}

\begin{table*}
    \centering
    \begin{tabular}{|c|c|p{50pt}p{45pt}p{40pt}|p{45pt}p{48pt}p{45pt}p{40pt}|}
    \bottomrule \hline
        \multirow{2}{*}{Method} & \multirow{2}{*}{Input} & \multicolumn{3}{c|}{SURREAL}  & \multicolumn{4}{c|}{PHSPD} \\
        \cline{3-9}
        & & \makecell[c]{PEL-MPJPE $\downarrow$} & \makecell[c]{PA-MPJPE $\downarrow$} & \makecell[c]{PCK $\uparrow$} & \makecell[c]{MPJPE $\downarrow$} & \makecell[c]{PEL-MPJPE $\downarrow$} & \makecell[c]{PA-MPJPE $\downarrow$} & \makecell[c]{PCK $\uparrow$} \\
    \hline
        \multirow{3}{*}{\makecell[c]{HumanSfP1\\\cite{zou2020detailed}}} & \makecell[c]{Polar} & \makecell[c]{99.33} & \makecell[c]{70.19} & \makecell[c]{0.65} & \makecell[c]{85.74} & \makecell[c]{83.70} & \makecell[c]{54.90} & \makecell[c]{0.72} \\
        & \makecell[c]{Polar+Mask} & \makecell[c]{95.28~(4.05)} & \makecell[c]{67.84~(2.35)} & \makecell[c]{0.67~(0.02)} & \makecell[c]{75.76~(9.98)} & \makecell[c]{73.04~(10.66)} & \makecell[c]{50.59~(4.31)} & \makecell[c]{0.77~(0.05)} \\
        & \makecell[c]{Polar+Normal} & \makecell[c]{84.78~(\textbf{14.55})} & \makecell[c]{60.82~(\textbf{9.37})} & \makecell[c]{0.72~(\textbf{0.07})} & \makecell[c]{66.97~(\textbf{18.77})} & \makecell[c]{63.12~(\textbf{20.58})} & \makecell[c]{42.08~(\textbf{12.82})} & \makecell[c]{0.83~(\textbf{0.11})} \\
    \hline
        \multirow{3}{*}{\makecell[c]{HumanSfP2\\(color)}} & \makecell[c]{Color} & \makecell[c]{96.27} & \makecell[c]{74.88} & \makecell[c]{0.69} & \makecell[c]{88.55} & \makecell[c]{77.97} & \makecell[c]{54.32} & \makecell[c]{0.74} \\
        & \makecell[c]{Color+Mask} & \makecell[c]{84.84~(11.43)} & \makecell[c]{62.74~(12.14)} & \makecell[c]{0.73~(0.04)} & \makecell[c]{80.79~(7.76)} & \makecell[c]{71.58~(6.39)} & \makecell[c]{44.55~(9.77)} & \makecell[c]{0.80~(0.06)} \\
        & \makecell[c]{Color+Normal} & \makecell[c]{80.28~(\textbf{15.99})} & \makecell[c]{59.39~(\textbf{15.49})} & \makecell[c]{0.76~(\textbf{0.07})} & \makecell[c]{76.34~(\textbf{12.21})} & \makecell[c]{68.23~(\textbf{9.74})} & \makecell[c]{41.95~(\textbf{12.37})} & \makecell[c]{0.84~(\textbf{0.10})} \\
    \hline
        \multirow{3}{*}{\makecell[c]{HumanSfP2\\(ours)}} & \makecell[c]{Polar} & \makecell[c]{74.12} & \makecell[c]{56.39} & \makecell[c]{0.78} & \makecell[c]{70.73} & \makecell[c]{63.82} & \makecell[c]{44.05} & \makecell[c]{0.83} \\
        & \makecell[c]{Polar+Mask} & \makecell[c]{72.20~(1.92)} & \makecell[c]{55.48~(0.91)} & \makecell[c]{0.78~(0)} & \makecell[c]{68.95~(1.78)} & \makecell[c]{60.28~(3.54)} & \makecell[c]{41.58~(2.47)} & \makecell[c]{0.85~(0.02)} \\
        & \makecell[c]{Polar+Normal} & \makecell[c]{59.17~(\textbf{14.95})} & \makecell[c]{46.58~(\textbf{9.81})} & \makecell[c]{0.85~(\textbf{0.07})} & \makecell[c]{62.04~(\textbf{8.69})} & \makecell[c]{54.98~(\textbf{8.84})} & \makecell[c]{36.88~(\textbf{7.17})} & \makecell[c]{0.88~(\textbf{0.05})} \\
    \hline \toprule 
    \end{tabular}
    \caption{Ablation study of hybrid input in human pose estimation on both SURREAL and PHSPD datasets. The unit of joint error is in millimeter. The number in round bracket shows the improvement over the case of polar/RGB as the sole input.}
    \label{tab:pose-estimation-ablation}
\end{table*}

Considering there is no previous work in this new task of single polarization image based clothed human shape reconstruction, four RGB-based methods are recruited for comparison. They are DepthHuman~\cite{tang2019neural}, HMD~\cite{zhu2021detailed}, PIFu~\cite{saito2019pifu}, PIFuHD~\cite{saito2020pifuhd} and PaMIR~\cite{zheng2021pamir}, where RGB image having closest view with polarization image is used as the input for evaluation. The estimated clothed shape is then compared with human point cloud to calculate 3D point to surface error (P2S).

Quantitative results are displayed in Table~\ref{tab:shape-estimation}. Here, DepthHuman performs the worst, which may be partly attribute to its consideration of the surface depth instead of the entire body shape. PIFu~\cite{saito2019pifu} and PIFuHD~\cite{saito2020pifuhd} show similar results, with PIFuHD having slightly larger error. The explanation is PIFu requires human mask as a prior, while PIFuHD do not have such assumption. However, both methods do not take human pose into consideration when predicting their implicit surface functions. The 3D error from \emph{HMD}~\cite{zhu2021detailed} is relatively small which may due to the accurate initial shape estimation. Our SfP approach achieves the best performance, which should be credit to its exploitation of the estimated normal maps.

Exemplar visual results are presented in Fig.~\ref{fig:detailed-shape-demo}. It is observed that for DepthHuman, only a partial mesh with respect to the view in the input image is produced, as is also evidenced in Table~\ref{tab:shape-estimation}. HMD, on the other hand, does not work well, as evidenced by the often error-prone surface details. This may be attributed to the less reliable shading representation, given the new environmental lighting and texture ambiguities existed in these RGB images. PIFu~\cite{saito2019pifu} and PIFuHD~\cite{saito2020pifuhd} are able to predict human with clothing details, but both suffer from the relative inaccurate pose inference, making their results ill-aligned with the subject in the image space. PaMIR~\cite{zheng2021pamir} estimates reasonable poses but reconstructs in-accurate human shapes, which could attribute to the in-correct information provided by their deep implicit function. Our HumanSfP approach is shown capable of producing reliable prediction of clothed body shapes, which again demonstrates the applicability of polarization imaging in shape estimation, as well as the benefit of engaging the surface normal maps in our approach.

\begin{figure}
    \centering
    \includegraphics[width=\columnwidth]{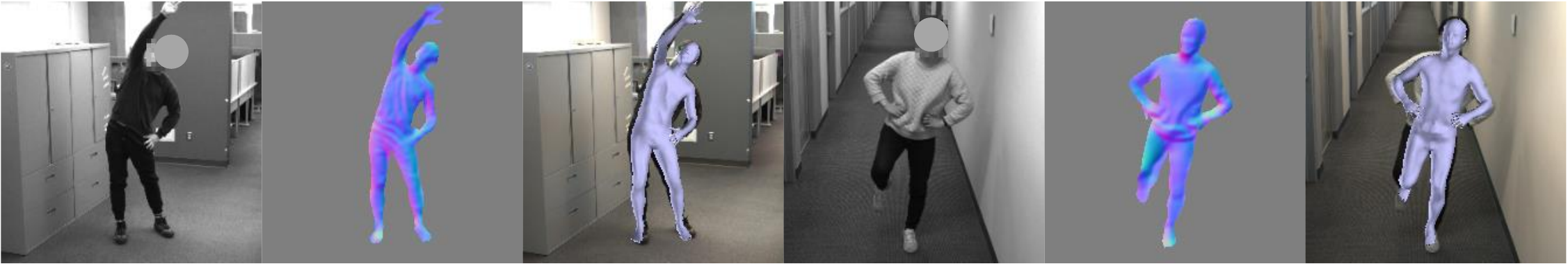}
    \caption{Exemplar estimation results of clothed body shapes, obtained on polarization images from new scene context.}
    \label{fig:extra-demo}
\end{figure}

Qualitative results presented in Fig.~\ref{fig:extra-demo} showcase the generalization ability of our approach. Note the polarization images are acquired at different physical locations with distinct background scenes that are very dissimilar to those in the training images. 

\subsection{Ablation Study}
\label{sec:ablation}

\textbf{Polarization and RGB modalities for normal estimation.} Here we want to compare the performance of normal map reconstruction from RGB images vs. polarization images. Let HumanSfP2~(color) denote the model that uses only RGB image, HumanSfP2~(no-prior) be the model without incorporating the ambiguous normal maps as the physical priors and with only the polarization image as input. The quantitative results are summarized in Table~\ref{tab:normal-estimation-ablation}. We observe that HumanSfP2~(ours) exceeds HumanSfP2~(color) by $\sim 7^{\circ}$ in SURREAL dataset and $\sim 3.5^{\circ}$ in PHSPD dataset. The larger improvement of MAE in SURREAL dataset may due to the fact that SURREAL dataset is synthesized from naked and minimal dressed body shapes, such that the normal map of human body is smooth thus relatively easier to predict than the real-world PHSPD dataset. Moreover, similar MAE results are presented by HumanSfP2~(no-prior) and HumanSfP2~(color), which showcases the importance of ambiguous normal maps in our approach that carries the critical geometric clues for high performance in surface normal estimation from polarization images.

\textbf{Normal maps for pose estimation.} This section demonstrates the significant performance gain that normal maps provides in pose estimation. As in Table~\ref{tab:pose-estimation-ablation}, the three methods of HumanSfP1\cite{zou2020detailed}, HumanSfP2~(color) and HumanSfP2~(ours) are equipped with different input combinations: polar/RGB image, polar/RGB image with foreground mask, and polar/RGB image with predicted normal map. Within each method, the performance gain is particularly significant when normal map is incorporated as input. This may due to the rich geometric information encoded in the normal map representation. Less significant gain is obtained when only mask is incorporated as input, which further demonstrates the effectiveness of geometric information in pose estimation. When comparing across polarization and RGB modalities in HumanSfP2~(ours) and HumanSfP2~(color), there is still noticeable improvement in HumanSfP2~(color), which combines RGB image with normal map as the input. However, the overall performance of HumanSfP2~(color) is worse than that of HumanSfP2~(ours). The explanation is that the normal maps estimated from RGB images are not as reliable as those obtained from the polarization image counterparts.

\section{Conclusion}
We tackle in this paper a new problem of estimating human shapes from single 2D polarization images. Our work exemplifies the applicability of engaging polarization cameras as a promising alternative to the existing imaging modalities for human pose and shape estimation. Moreover, by exploiting the rich geometric details in the surface normal of the input polarization images, our SfP approach is capable of reconstructing human body shapes of surface details. We expect this could be a useful tool in many downstream applications.


%




\ifCLASSOPTIONcaptionsoff
  \newpage
\fi



\bibliographystyle{IEEEtran}
\bibliography{IEEEabrv}

\begin{thebibliography}{10}
\providecommand{\url}[1]{#1}
\csname url@samestyle\endcsname
\providecommand{\newblock}{\relax}
\providecommand{\bibinfo}[2]{#2}
\providecommand{\BIBentrySTDinterwordspacing}{\spaceskip=0pt\relax}
\providecommand{\BIBentryALTinterwordstretchfactor}{4}
\providecommand{\BIBentryALTinterwordspacing}{\spaceskip=\fontdimen2\font plus
\BIBentryALTinterwordstretchfactor\fontdimen3\font minus
  \fontdimen4\font\relax}
\providecommand{\BIBforeignlanguage}[2]{{%
\expandafter\ifx\csname l@#1\endcsname\relax
\typeout{** WARNING: IEEEtran.bst: No hyphenation pattern has been}%
\typeout{** loaded for the language `#1'. Using the pattern for}%
\typeout{** the default language instead.}%
\else
\language=\csname l@#1\endcsname
\fi
#2}}
\providecommand{\BIBdecl}{\relax}
\BIBdecl

\bibitem{ning2017knowledge}
G.~Ning, Z.~Zhang, and Z.~He, ``Knowledge-guided deep fractal neural networks
  for human pose estimation,'' \emph{IEEE Transactions on Multimedia}, vol.~20,
  no.~5, pp. 1246--1259, 2017.

\bibitem{kamel2020hybrid}
A.~Kamel, B.~Sheng, P.~Li, J.~Kim, and D.~D. Feng, ``Hybrid
  refinement-correction heatmaps for human pose estimation,'' \emph{IEEE
  Transactions on Multimedia}, vol.~23, pp. 1330--1342, 2020.

\bibitem{zhang2021single}
Q.~Zhang, Y.~Jiang, Q.~Zhou, Y.~Zhao, Y.~Liu, H.~Lu, and X.-S. Hua, ``Single
  person dense pose estimation via geometric equivariance consistency,''
  \emph{IEEE Transactions on Multimedia}, 2021.

\bibitem{park20163d}
S.~Park, J.~Hwang, and N.~Kwak, ``3d human pose estimation using convolutional
  neural networks with 2d pose information,'' in \emph{ECCV}, 2016.

\bibitem{zhao2017simple}
R.~Zhao, Y.~Wang, and A.~M. Martinez, ``A simple, fast and highly-accurate
  algorithm to recover 3d shape from 2d landmarks on a single image,''
  \emph{IEEE Transactions on Pattern Analysis and Machine Intelligence},
  vol.~40, no.~12, pp. 3059--3066, 2017.

\bibitem{nie2017monocular}
B.~X. Nie, P.~Wei, and S.-C. Zhu, ``Monocular 3d human pose estimation by
  predicting depth on joints,'' in \emph{ICCV}, 2017.

\bibitem{wang2018drpose3d}
M.~Wang, X.~Chen, W.~Liu, C.~Qian, L.~Lin, and L.~Ma, ``Drpose3d: depth ranking
  in 3d human pose estimation,'' in \emph{IJCAI}, 2018.

\bibitem{fang2018learning}
H.-S. Fang, Y.~Xu, W.~Wang, X.~Liu, and S.-C. Zhu, ``Learning pose grammar to
  encode human body configuration for 3d pose estimation,'' in \emph{AAAI},
  2018.

\bibitem{habibie2019wild}
I.~Habibie, W.~Xu, D.~Mehta, G.~Pons-Moll, and C.~Theobalt, ``In the wild human
  pose estimation using explicit 2d features and intermediate 3d
  representations,'' in \emph{CVPR}, 2019.

\bibitem{wang20193d}
K.~Wang, L.~Lin, C.~Jiang, C.~Qian, and P.~Wei, ``3d human pose machines with
  self-supervised learning,'' \emph{IEEE Transactions on Pattern Analysis and
  Machine Intelligence}, vol.~42, no.~5, pp. 1069--1082, 2020.

\bibitem{anguelov2005scape}
D.~Anguelov, P.~Srinivasan, D.~Koller, S.~Thrun, J.~Rodgers, and J.~Davis,
  ``Scape: shape completion and animation of people,'' in \emph{ACM
  transactions on graphics}, vol.~24, no.~3.\hskip 1em plus 0.5em minus
  0.4em\relax ACM, 2005, pp. 408--416.

\bibitem{loper2015smpl}
M.~Loper, N.~Mahmood, J.~Romero, G.~Pons-Moll, and M.~J. Black, ``Smpl: A
  skinned multi-person linear model,'' \emph{ACM transactions on graphics},
  vol.~34, no.~6, p. 248, 2015.

\bibitem{zhao20183}
T.~Zhao, S.~Li, K.~N. Ngan, and F.~Wu, ``3-d reconstruction of human body shape
  from a single commodity depth camera,'' \emph{IEEE Transactions on
  Multimedia}, vol.~21, no.~1, pp. 114--123, 2018.

\bibitem{lu2020parametric}
Y.~Lu, J.-H. Cha, S.-K. Youm, and S.-W. Jung, ``Parametric shape estimation of
  human body under wide clothing,'' \emph{IEEE Transactions on Multimedia},
  2020.

\bibitem{zuo2020sparsefusion}
X.~Zuo, S.~Wang, J.~Zheng, W.~Yu, M.~Gong, R.~Yang, and L.~Cheng,
  ``Sparsefusion: Dynamic human avatar modeling from sparse rgbd images,''
  \emph{IEEE Transactions on Multimedia}, vol.~23, pp. 1617--1629, 2020.

\bibitem{balan2007detailed}
A.~O. Balan, L.~Sigal, M.~J. Black, J.~E. Davis, and H.~W. Haussecker,
  ``Detailed human shape and pose from images,'' in \emph{CVPR}, 2007.

\bibitem{dibra2016hs}
E.~Dibra, H.~Jain, C.~{\"O}ztireli, R.~Ziegler, and M.~Gross, ``Hs-nets:
  Estimating human body shape from silhouettes with convolutional neural
  networks,'' in \emph{3DV}, 2016.

\bibitem{bogo2016keep}
F.~Bogo, A.~Kanazawa, C.~Lassner, P.~Gehler, J.~Romero, and M.~J. Black, ``Keep
  it smpl: Automatic estimation of 3d human pose and shape from a single
  image,'' in \emph{ECCV}, 2016.

\bibitem{lassner2017unite}
C.~Lassner, J.~Romero, M.~Kiefel, F.~Bogo, M.~J. Black, and P.~V. Gehler,
  ``Unite the people: Closing the loop between 3d and 2d human
  representations,'' in \emph{CVPR}, 2017.

\bibitem{kanazawa2018end}
A.~Kanazawa, M.~J. Black, D.~W. Jacobs, and J.~Malik, ``End-to-end recovery of
  human shape and pose,'' in \emph{CVPR}, 2018.

\bibitem{pavlakos2018learning}
G.~Pavlakos, L.~Zhu, X.~Zhou, and K.~Daniilidis, ``Learning to estimate 3d
  human pose and shape from a single color image,'' in \emph{CVPR}, 2018.

\bibitem{kanazawa2019learning}
A.~Kanazawa, J.~Y. Zhang, P.~Felsen, and J.~Malik, ``Learning 3d human dynamics
  from video,'' in \emph{CVPR}, 2019.

\bibitem{kocabas2020vibe}
M.~Kocabas, N.~Athanasiou, and M.~J. Black, ``Vibe: Video inference for human
  body pose and shape estimation,'' in \emph{CVPR}, 2020.

\bibitem{zheng2021pamir}
Z.~Zheng, T.~Yu, Y.~Liu, and Q.~Dai, ``Pamir: Parametric model-conditioned
  implicit representation for image-based human reconstruction,'' \emph{IEEE
  Transactions on Pattern Analysis and Machine Intelligence}, 2021.

\bibitem{zhu2021detailed}
H.~Zhu, X.~Zuo, H.~Yang, S.~Wang, X.~Cao, and R.~Yang, ``Detailed avatar
  recovery from single image,'' \emph{IEEE Transactions on Pattern Analysis and
  Machine Intelligence}, 2021.

\bibitem{yang2018polarimetric}
L.~Yang, F.~Tan, A.~Li, Z.~Cui, Y.~Furukawa, and P.~Tan, ``Polarimetric dense
  monocular slam,'' in \emph{CVPR}, 2018, pp. 3857--3866.

\bibitem{ba2019physics}
Y.~Ba, A.~Gilbert, F.~Wang, J.~Yang, R.~Chen, Y.~Wang, L.~Yan, B.~Shi, and
  A.~Kadambi, ``Deep shape from polarization,'' in \emph{ECCV}, 2020.

\bibitem{wehner2006significance}
R.~Wehner and M.~M{\"u}ller, ``The significance of direct sunlight and
  polarized skylight in the ant’s celestial system of navigation,''
  \emph{Proceedings of the National Academy of Sciences}, vol. 103, no.~33, pp.
  12\,575--12\,579, 2006.

\bibitem{daly2016dynamic}
I.~M. Daly, M.~J. How, J.~C. Partridge, S.~E. Temple, N.~J. Marshall, T.~W.
  Cronin, and N.~W. Roberts, ``Dynamic polarization vision in mantis shrimps,''
  \emph{Nature communications}, vol.~7, p. 12140, 2016.

\bibitem{tang2019neural}
S.~Tang, F.~Tan, K.~Cheng, Z.~Li, S.~Zhu, and P.~Tan, ``A neural network for
  detailed human depth estimation from a single image,'' in \emph{ICCV}, 2019.

\bibitem{zou2020detailed}
S.~Zou, X.~Zuo, Y.~Qian, S.~Wang, C.~Xu, M.~Gong, and L.~Cheng, ``3d human
  shape reconstruction from a polarization image,'' in \emph{ECCV}, 2020.

\bibitem{human36m}
C.~Ionescu, D.~Papava, V.~Olaru, and C.~Sminchisescu, ``Human3.6m: Large scale
  datasets and predictive methods for 3d human sensing in natural
  environments,'' \emph{IEEE Transactions on Pattern Analysis and Machine
  Intelligence}, vol.~36, no.~7, pp. 1325--1339, 2014.

\bibitem{mehta2017monocular}
D.~Mehta, H.~Rhodin, D.~Casas, P.~Fua, O.~Sotnychenko, W.~Xu, and C.~Theobalt,
  ``Monocular 3d human pose estimation in the wild using improved cnn
  supervision,'' in \emph{3DV}, 2017.

\bibitem{atkinson2006recovery}
G.~A. Atkinson and E.~R. Hancock, ``Recovery of surface orientation from
  diffuse polarization,'' \emph{IEEE Transactions on Image Processing},
  vol.~15, no.~6, pp. 1653--1664, 2006.

\bibitem{kadambi2017depth}
A.~Kadambi, V.~Taamazyan, B.~Shi, and R.~Raskar, ``Depth sensing using
  geometrically constrained polarization normals,'' \emph{International Journal
  of Computer Vision}, vol. 125, no. 1-3, pp. 34--51, 2017.

\bibitem{chen2018polarimetric}
L.~Chen, Y.~Zheng, A.~Subpa-Asa, and I.~Sato, ``Polarimetric three-view
  geometry,'' in \emph{ECCV}, 2018.

\bibitem{cui2017polarimetric}
Z.~Cui, J.~Gu, B.~Shi, P.~Tan, and J.~Kautz, ``Polarimetric multi-view
  stereo,'' in \emph{CVPR}, 2017.

\bibitem{zhou2016sparseness}
X.~Zhou, M.~Zhu, S.~Leonardos, K.~G. Derpanis, and K.~Daniilidis, ``Sparseness
  meets deepness: 3d human pose estimation from monocular video,'' in
  \emph{CVPR}, 2016.

\bibitem{akhter2015pose}
I.~Akhter and M.~J. Black, ``Pose-conditioned joint angle limits for 3d human
  pose reconstruction,'' in \emph{CVPR}, 2015, pp. 1446--1455.

\bibitem{wang2014robust}
C.~Wang, Y.~Wang, Z.~Lin, A.~L. Yuille, and W.~Gao, ``Robust estimation of 3d
  human poses from a single image,'' in \emph{CVPR}, 2014.

\bibitem{li2015maximum}
S.~Li, W.~Zhang, and A.~B. Chan, ``Maximum-margin structured learning with deep
  networks for 3d human pose estimation,'' in \emph{ICCV}, 2015.

\bibitem{ci2019optimizing}
H.~Ci, C.~Wang, X.~Ma, and Y.~Wang, ``Optimizing network structure for 3d human
  pose estimation,'' in \emph{ICCV}, 2019.

\bibitem{cai2019exploiting}
Y.~Cai, L.~Ge, J.~Liu, J.~Cai, T.-J. Cham, J.~Yuan, and N.~M. Thalmann,
  ``Exploiting spatial-temporal relationships for 3d pose estimation via graph
  convolutional networks,'' in \emph{ICCV}, 2019.

\bibitem{tome2017lifting}
D.~Tome, C.~Russell, and L.~Agapito, ``Lifting from the deep: convolutional 3d
  pose estimation from a single image,'' in \emph{CVPR}, 2017.

\bibitem{martinez2017simple}
J.~Martinez, R.~Hossain, J.~Romero, and J.~J. Little, ``A simple yet effective
  baseline for 3d human pose estimation,'' in \emph{ICCV}, 2017.

\bibitem{zhou2017towards}
X.~Zhou, Q.~Huang, X.~Sun, X.~Xue, and Y.~Wei, ``Towards 3d human pose
  estimation in the wild: a weakly-supervised approach,'' in \emph{ICCV}, 2017.

\bibitem{yang20183d}
W.~Yang, W.~Ouyang, X.~Wang, J.~Ren, H.~Li, and X.~Wang, ``3d human pose
  estimation in the wild by adversarial learning,'' in \emph{CVPR}, 2018.

\bibitem{pavlakos2018ordinal}
G.~Pavlakos, X.~Zhou, and K.~Daniilidis, ``Ordinal depth supervision for 3d
  human pose estimation,'' in \emph{CVPR}, 2018.

\bibitem{wandt2019repnet}
B.~Wandt and B.~Rosenhahn, ``Repnet: weakly supervised training of an
  adversarial reprojection network for 3d human pose estimation,'' in
  \emph{CVPR}, 2019.

\bibitem{dibra2017human}
E.~Dibra, H.~Jain, C.~Oztireli, R.~Ziegler, and M.~Gross, ``Human shape from
  silhouettes using generative hks descriptors and cross-modal neural
  networks,'' in \emph{CVPR}, 2017.

\bibitem{varol2017learning}
G.~Varol, J.~Romero, X.~Martin, N.~Mahmood, M.~J. Black, I.~Laptev, and
  C.~Schmid, ``Learning from synthetic humans,'' in \emph{CVPR}, 2017.

\bibitem{omran2018neural}
M.~Omran, C.~Lassner, G.~Pons-Moll, P.~Gehler, and B.~Schiele, ``Neural body
  fitting: Unifying deep learning and model based human pose and shape
  estimation,'' in \emph{3DV}, 2018.

\bibitem{xu2019denserac}
Y.~Xu, S.-C. Zhu, and T.~Tung, ``Denserac: Joint 3d pose and shape estimation
  by dense render-and-compare,'' in \emph{ICCV}, 2019.

\bibitem{kolotouros2019learning}
N.~Kolotouros, G.~Pavlakos, M.~J. Black, and K.~Daniilidis, ``Learning to
  reconstruct 3d human pose and shape via model-fitting in the loop,'' in
  \emph{CVPR}, 2019.

\bibitem{jiang2019skeleton}
H.~Jiang, J.~Cai, and J.~Zheng, ``Skeleton-aware 3d human shape reconstruction
  from point clouds,'' in \emph{ICCV}, 2019.

\bibitem{varol2018bodynet}
G.~Varol, D.~Ceylan, B.~Russell, J.~Yang, E.~Yumer, I.~Laptev, and C.~Schmid,
  ``Bodynet: volumetric inference of 3d human body shapes,'' in \emph{ECCV},
  2018.

\bibitem{zheng2019deephuman}
Z.~Zheng, T.~Yu, Y.~Wei, Q.~Dai, and Y.~Liu, ``Deephuman: 3d human
  reconstruction from a single image,'' in \emph{ICCV}, 2019.

\bibitem{saito2019pifu}
S.~Saito, Z.~Huang, R.~Natsume, S.~Morishima, A.~Kanazawa, and H.~Li, ``Pifu:
  pixel-aligned implicit function for high-resolution clothed human
  digitization,'' in \emph{ICCV}, 2019.

\bibitem{yang2021s3}
Z.~Yang, S.~Wang, S.~Manivasagam, Z.~Huang, W.-C. Ma, X.~Yan, E.~Yumer, and
  R.~Urtasun, ``S3: Neural shape, skeleton, and skinning fields for 3d human
  modeling,'' in \emph{CVPR}, 2021.

\bibitem{saito2020pifuhd}
S.~Saito, T.~Simon, J.~Saragih, and H.~Joo, ``Pifuhd: Multi-level pixel-aligned
  implicit function for high-resolution 3d human digitization,'' in
  \emph{CVPR}, 2020.

\bibitem{lin2014microsoft}
T.-Y. Lin, M.~Maire, S.~Belongie, J.~Hays, P.~Perona, D.~Ramanan,
  P.~Doll{\'a}r, and C.~L. Zitnick, ``Microsoft coco: Common objects in
  context,'' in \emph{ECCV}, 2014.

\bibitem{andriluka2018posetrack}
M.~Andriluka, U.~Iqbal, E.~Insafutdinov, L.~Pishchulin, A.~Milan, J.~Gall, and
  B.~Schiele, ``Posetrack: A benchmark for human pose estimation and
  tracking,'' in \emph{CVPR}, 2018.

\bibitem{von2018recovering}
T.~von Marcard, R.~Henschel, M.~J. Black, B.~Rosenhahn, and G.~Pons-Moll,
  ``Recovering accurate 3d human pose in the wild using imus and a moving
  camera,'' in \emph{ECCV}, 2018.

\bibitem{murray1994mathematical}
R.~M. Murray, Z.~Li, S.~S. Sastry, and S.~S. Sastry, \emph{A mathematical
  introduction to robotic manipulation}.\hskip 1em plus 0.5em minus 0.4em\relax
  CRC press, 1994.

\bibitem{nehab2005efficiently}
D.~Nehab, S.~Rusinkiewicz, J.~Davis, and R.~Ramamoorthi, ``Efficiently
  combining positions and normals for precise 3d geometry,'' \emph{ACM
  transactions on graphics}, vol.~24, no.~3, pp. 536--543, 2005.

\bibitem{openpose}
Z.~Cao, G.~H. Martinez, T.~Simon, S.~Wei, and Y.~A. Sheikh, ``Openpose:
  Realtime multi-person 2d pose estimation using part affinity fields,''
  \emph{IEEE Transactions on Pattern Analysis and Machine Intelligence},
  vol.~43, no.~1, pp. 172--186, 2021.

\bibitem{pavlakos2019smplifyx}
G.~Pavlakos, V.~Choutas, N.~Ghorbani, T.~Bolkart, A.~A. Osman, D.~Tzionas, and
  M.~J. Black, ``Expressive body capture: 3d hands, face, and body from a
  single image,'' in \emph{CVPR}, 2019.

\bibitem{bollapragada2018LBFGS}
R.~Bollapragada, J.~Nocedal, D.~Mudigere, H.-J. Shi, and P.~T.~P. Tang, ``A
  progressive batching l-bfgs method for machine learning,'' in \emph{ICML},
  2018.

\bibitem{kingma2014adam}
D.~P. Kingma and J.~Ba, ``Adam: {A} method for stochastic optimization,'' in
  \emph{ICLR}, 2015.

\bibitem{he2016deep}
K.~He, X.~Zhang, S.~Ren, and J.~Sun, ``Deep residual learning for image
  recognition,'' in \emph{CVPR}, 2016.

\bibitem{smith2016linear}
W.~A. Smith, R.~Ramamoorthi, and S.~Tozza, ``Height-from-polarisation with
  unknown lighting or albedo,'' \emph{IEEE Transactions on Pattern Analysis and
  Machine Intelligence}, vol.~41, no.~12, pp. 2875--2888, 2018.

\end{thebibliography}
%



%




\begin{IEEEbiography}
	[{\includegraphics[width=1in,height=1.25in,clip,keepaspectratio]{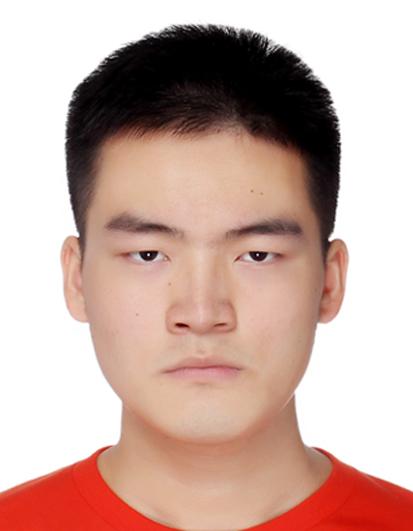}}]
	{Shihao Zou} received the B.Sc. degree from Beijing Institute of Technology, China, in 2017, and the M.Res. degree from University College London, UK, in 2018. He is currently a Ph.D. candidate at University of Alberta. His interests include computer vision and machine learning, especially human pose and shape estimation, motion capture system.
\end{IEEEbiography}

\begin{IEEEbiography}
	[{\includegraphics[width=1in,height=1.25in,clip,keepaspectratio]{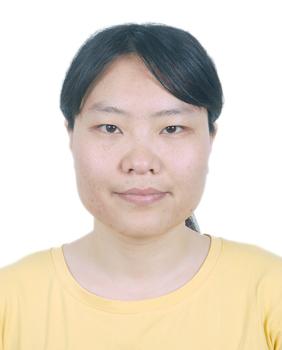}}]
	{Xinxin Zuo} received the M.E. degree from Northwestern Polytechnical University and Ph.D. degree from the University of Kentucky. She is currently a Postdoctoral Fellow at University of Alberta. Her interests include computer vision and graphics, especially on 3D reconstruction and human modeling.
\end{IEEEbiography}

\begin{IEEEbiography}
	[{\includegraphics[width=1in,height=1.25in,clip,keepaspectratio]{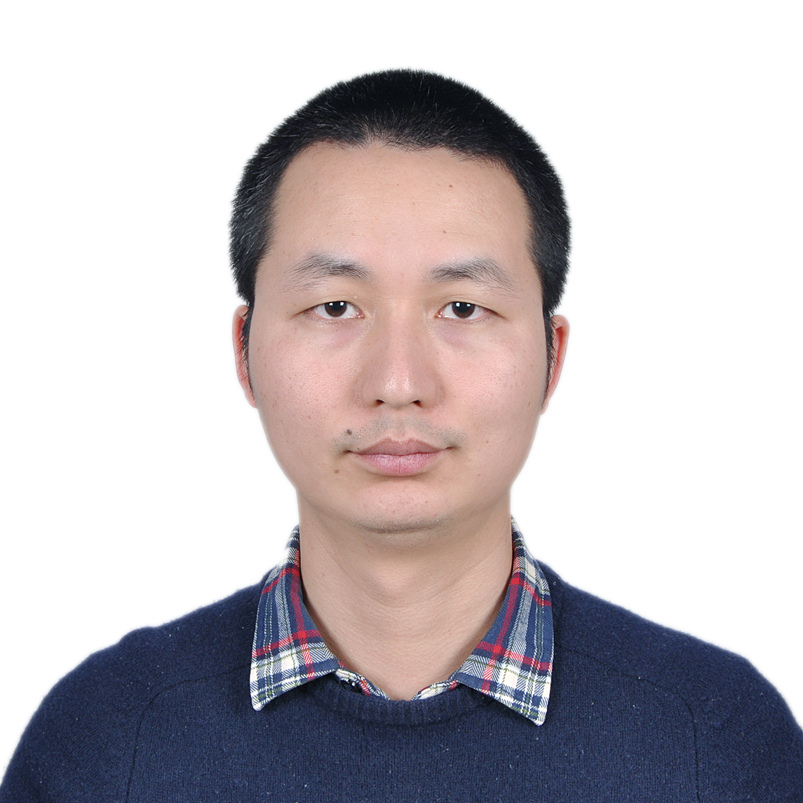}}]
	{Sen Wang} received the B.E. degree and Ph.D. degree from Northwestern Polytechnical University. From 2015 to 2016, he was a Visiting Ph.D. Student at the University of Kentucky. He is currently a Postdoctoral Fellow at University of Alberta. His research interests include computer vision and robotics. 
\end{IEEEbiography}

\begin{IEEEbiography}
	[{\includegraphics[width=1in,height=1.25in,clip,keepaspectratio]{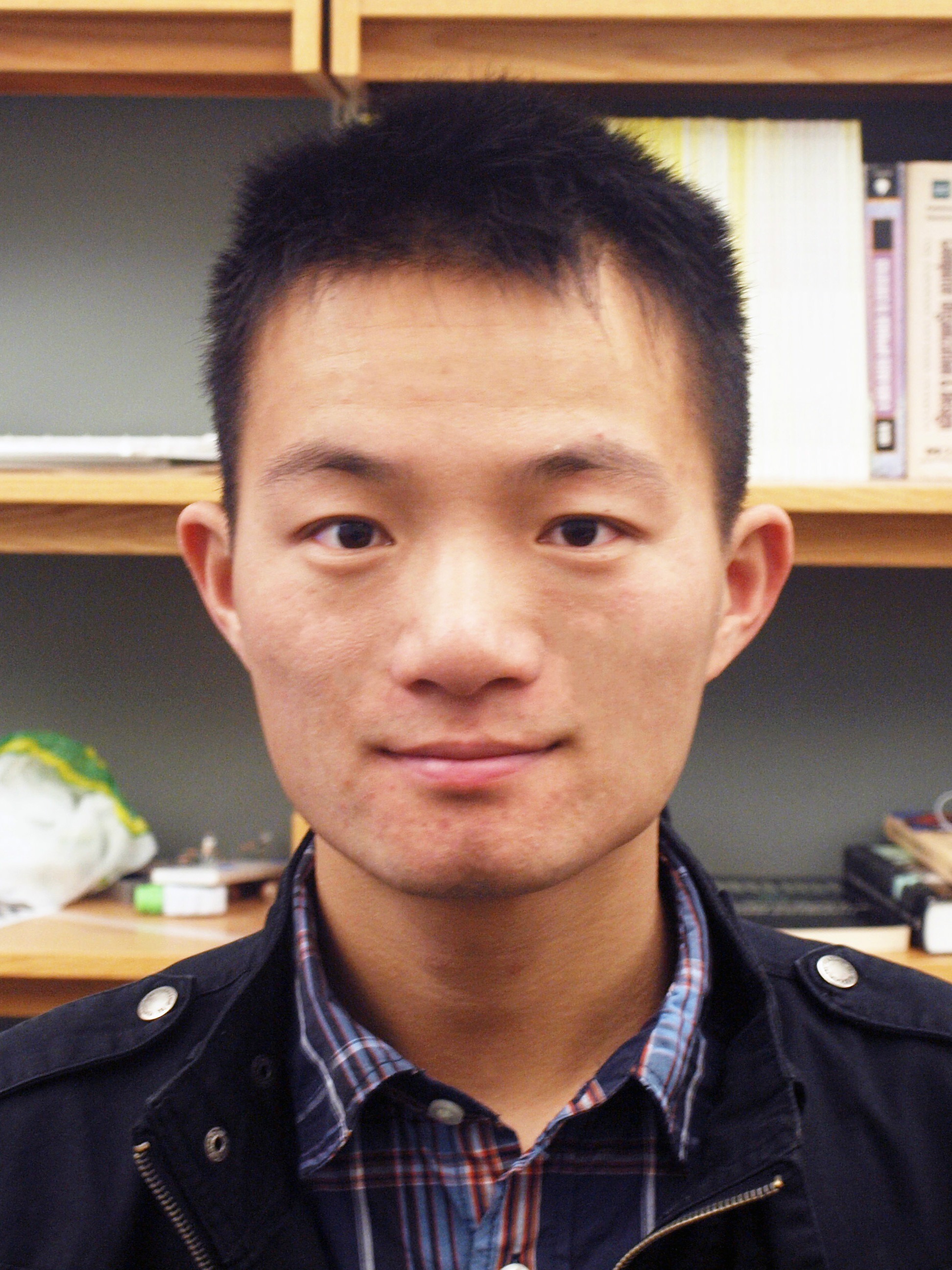}}]{Yiming Qian} received the B.Sc. degree from University of Science and Technology of China, Hefei, China, in 2012, the M.Sc. degree from Memorial University of Newfoundland, St. John’s, Canada, in 2014, and the Ph.D. degree from the Department of Computing Science, University of Alberta, Edmonton, Canada, in 2019. He is currently an Assistant Professor in the Department of Computer Science at the University of Manitoba, Winnipeg, Canada. His research interests include computer vision and graphics, while he recently focuses on 3D modeling.
\end{IEEEbiography}

\begin{IEEEbiography}
	[{\includegraphics[width=1in,height=1.25in,clip,keepaspectratio]{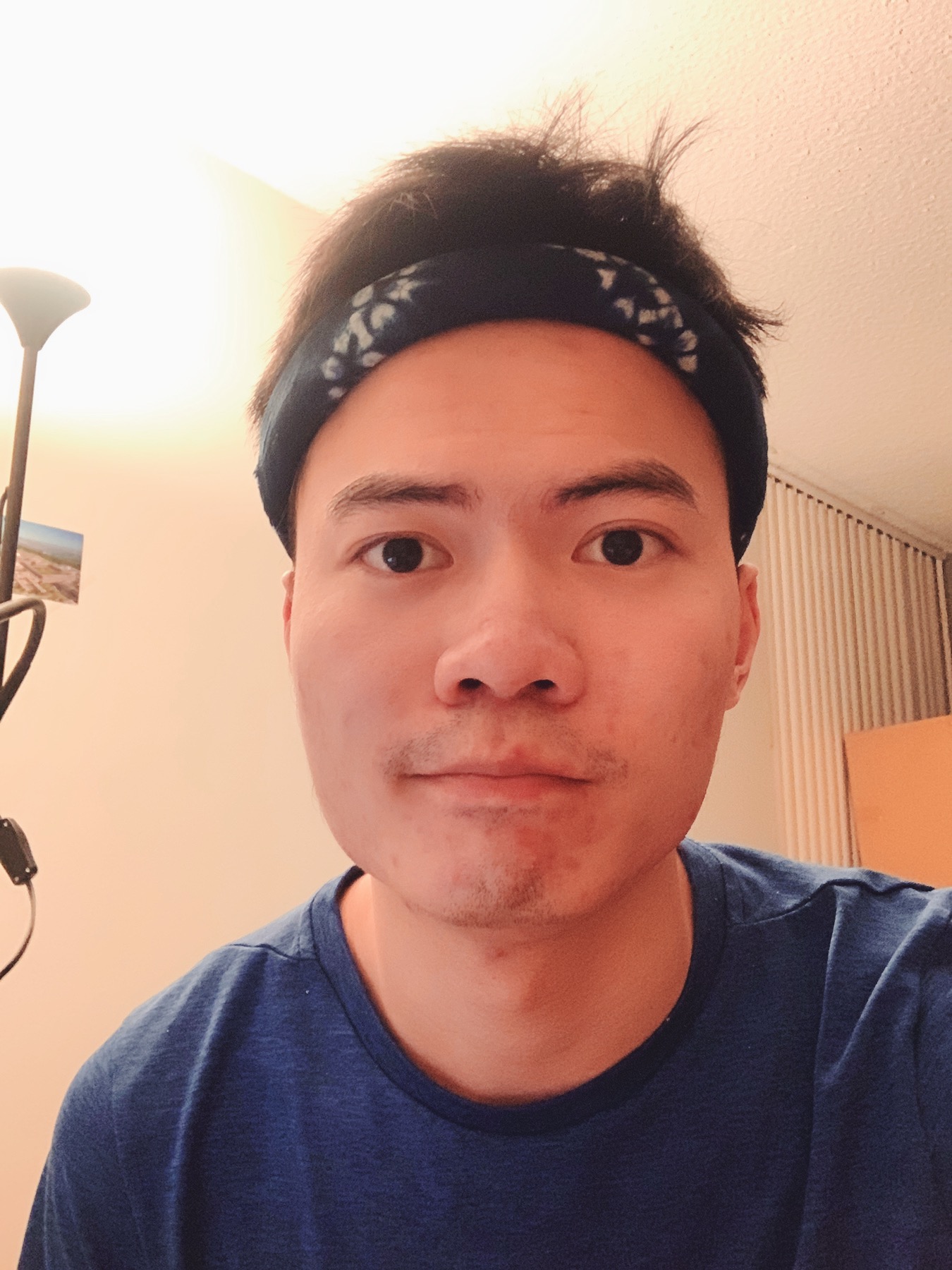}}]{Chuan Guo} obtained his bachelor's degree from Jilin University in 2017, and is currently a Ph.D. student in the Department of Electrical and Computer Engineering at the University of Alberta. His research interests mainly lie in computer vision and human motion modeling, such as multi-modal human motion generation. 
\end{IEEEbiography}

\begin{IEEEbiography}
	[{\includegraphics[width=1in,height=1.25in,clip,keepaspectratio]{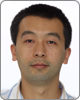}}]{Li Cheng} received the Ph.D. degree in computer science from the University of Alberta, Canada. He is an associate professor with the Department of Electrical and Computer Engineering, University of Alberta. Prior to coming back to University of Alberta, He has worked at A*STAR, Singapore, TTI-Chicago, USA, and NICTA, Australia. His research expertise is mainly on computer vision and machine learning. He is a senior member of the IEEE, and is serving as the editor of IEEE Transactions on Multimedia and Pattern Recognition.
\end{IEEEbiography}



\end{document}


%
\title{Supplementary Material\\Human Pose and Shape Estimation from Single Polarization Images}
%
%
%

\author{Shihao~Zou,~\IEEEmembership{Member,~IEEE},
        Xinxin~Zuo,
        Sen~Wang,
        Yiming~Qian, 
        Chuan~Guo,
        and~Li~Cheng~\IEEEmembership{Senior~Member,~IEEE}
\thanks{S. Zou, X. Zuo, S. Wang and C. Guo are with the Department
of Electrical and Computer Engineering, University of Alberta, AB, Canada. (E-mail: szou2@ualberta.ca, xzuo@ualberta.ca, sen9@ualberta.ca, cguo2@ualberta.ca, wji3@ualberta.ca, jingjin1@ualberta.ca).}
\thanks{Y. Qian is with the Department of Computer Science, University of Manitoba, MB , Canada. (E-mail: yiming.qian@umanitoba.ca)}
\thanks{L. Cheng is the Department
of Electrical and Computer Engineering, University of Alberta, AB,
Canada. (E-mail: lcheng5@ualberta.ca)}
\thanks{Manuscript received January 05 2022; accepted March 18 2022. Li Cheng is the corresponding author for this paper.}}

%
%

\markboth{Journal of \LaTeX\ Class Files,~Vol.~14, No.~8, August~2015}%
{Shell \MakeLowercase{\textit{et al.}}: Bare Demo of IEEEtran.cls for IEEE Journals}
%



\maketitle

%
\IEEEpeerreviewmaketitle

\section{Shape Reconstruction}
The initial human shape obtained by SMPL representation still lacks fine surface details. Therefore, the aim of this step is to refine the initial SMPL shape guided by our surface normal estimate, as follows. The SMPL body shape is rendered on the image plane to form an initial depth map. The technique of~\cite{nehab2005efficiently} is then engaged here to obtain an optimized depth map $I_d$ from the predicted surface normal $\mathbf{\hat m}$ and the initial depth $\hat I_d$ map by minimizing the objective function,
\begin{equation}
    E(I_d) = \lambda_n E_n(I_d) + \lambda_d E_d(I_d) + \lambda_s E_s(I_d),
    \label{eq:integrate}
\end{equation}
which contains three energy terms. The first term, $E_n(I_d)$, ensures the predicted normal to be perpendicular to the tangents of the optimized depth surface,
\begin{equation}
     E_n(I_d) = \sum_{i,j} T_{u[i, j]} \mathbf{\hat m}_{[i, j]} + T_{v[i,j]} \mathbf{\hat m}_{[i, j]}.
\end{equation}
Here $[i,j]$ denotes a pixel coordinate. $u$ and $v$ represent the horizontal and vertical direction of the image plane, respectively. The tangents $T_u$ and $T_v$ are defined as
\begin{align}
     T_u &= \left(\frac{1}{f_u} \left(\frac{\partial I_d}{\partial u}(u-p_u) + I_d \right), \frac{1}{f_v}\frac{\partial I_d}{\partial u}(v-p_v),  \frac{\partial I_d}{\partial u}\right)^{\top}, \\
     T_v &= \left( \frac{1}{f_u}\frac{\partial I_d}{\partial u}(v-p_v), 
     \frac{1}{f_v} \left(\frac{\partial I_d}{\partial v}(v-p_v) +I_d \right),
     \frac{\partial I_d}{\partial v}\right)^{\top},
\end{align}
where $f_u$ and $f_v$ denote the focal length, $p_u$ and $p_v$ define the camera center coordinate, respectively.
The second term, $E_d(I_d)$, encourages the optimized depth to be close to the initial depth,
\begin{equation}
     E_d(I_d) = \sum_{i,j} \Big[ \Big( (\frac{i-p_v}{f_v})^2 + (\frac{j-p_u}{f_u})^2 +1 \Big) \big(I_{d[i,j]} - \hat{I}_{d[i,j]} \big) \Big]^2.
\end{equation}
The third and final term preserves smoothness of nearby pixels over the optimized depth map,
\begin{equation}
     E_s(I_d) = \sum_{i,j}\sum_{[i',j']\in \mathcal{N}(i,j)} \left\| I_{d[i, j]}-I_{d[i',j']} \right\|^2.
\end{equation}
Our depth map estimate is therefore obtained as a solution of the above mentioned linear least-squares system.  
Finally, our clothed body shape is produced by upsampling \& deforming the SMPL mesh according to the Laplacian of the optimized depth map.

\section{Polarization Image Synthesis}
A polarization image (polarizers of $0^{\circ}$, $45^{\circ}$, $90^{\circ}$ and $135^{\circ}$) can be synthesized from the rendered depth and color image. In detail, from the depth image, we obtain the normal map and calculate the zenith and azimuth angle, and from the color image, we get the gray image and take it as the polarization image of $0^{\circ}$ degree polarizer, denoted by $I(0^{\circ})$. Assuming diffuse reflection of the human body surface, the degree of polarization $\rho$ can be calculated according to the equation,
\begin{equation}
    \label{eq:zenith-angle}
    \rho = \frac{(n-\frac{1}{n})^2\sin^2\theta}{2+2n^2-(n+\frac{1}{n})^2\sin^2\theta+4\cos\theta\sqrt{n^2-\sin^2\theta}},
\end{equation}
with the calculated zenith angle for each pixel and refractive index known as 1.5. Then the upper and lower bound of the illumination intensity $I_{max}$ and $I_{min}$ can be solved in closed-form with the constraints as follows,
\begin{equation}
    \label{eq:degree-of-polarization}
    \rho = \frac{I_{max} - I_{min}}{I_{max} + I_{min}},
\end{equation}
and
\begin{equation}
    \label{eq:polar}
    I(0)=\frac{I_{max} + I_{min}}{2} + \frac{I_{max} - I_{min}}{2} \cos (2\varphi)).
\end{equation}
Finally, we can use the equation, 
\begin{equation}
    \label{eq:polar}
    I(\phi_{pol})=\frac{I_{max} + I_{min}}{2} + \frac{I_{max} - I_{min}}{2} \cos (2(\phi_{pol}-\varphi)),
\end{equation}
to get the image for polarizer $\phi_{pol}$ of $45^{\circ}$, $90^{\circ}$ and $135^{\circ}$ , or any arbitrary degree. To make it close to the real-world applications, we add Gaussian noise with the standard deviation $\sigma=1/255$ to each pixel of the synthetic polarization image and then quantize the intensity value to 8 bits. Due to the fact that we only have geometric information for human body, the synthetic polarization images only have values on human body part.

\section{PHSPD Dataset}

\begin{figure}
    \centering
    \includegraphics[width=\columnwidth]{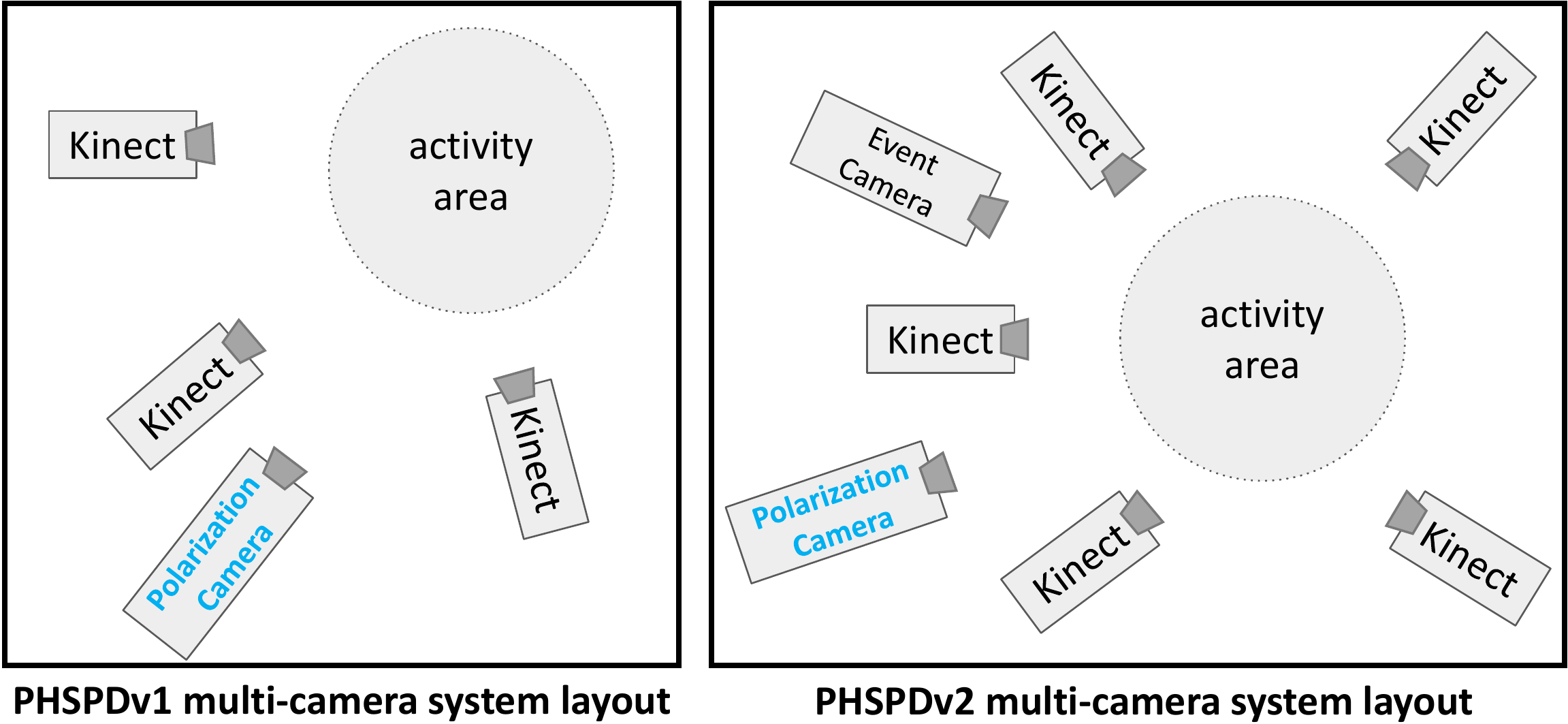}
    \caption{Left figure: the layout of multi-camera system in our PHSPDv1. Three RGB-D (Kinect V2) are placed around a circle of activity area with a polarization camera. Right figure: the layout of multi-camera system in our PHSPDv2. The number of RGB-D cameras is extended to five with a polarization camera and an event camera.}
    \label{fig:cam-layout}
\end{figure}

The layouts of our multi-camera acquisition system for PHSPDv1 \& v2 are shown in Fig.~\ref{fig:cam-layout}. In PHSPDv1, 7 cameras, three RGB-D cameras and a polarization camera, are synchronized, and in PHSPDv2, we extend the number of cameras in our system to be 12, five RGB-D cameras, a polarization camera and an event camera.

\textbf{Multi-camera Synchronization.} The multi-camera system in our PHSPD is mostly soft synchronized. In PHSPDv1, each camera is connected with a desktop, where the desktop with the polarization camera is the master and the other three ones with three Kinects V2 are clients. The master desktop uses TCP-IP protocol to communicate with the other clients. After receiving certain message, each client will copy the most recent frame data captured by the Kinect into the desktop memory. At the same time, the master desktop sends a software trigger to the polarization camera to capture one frame into the camera buffer. For PHSPDv2, three latest RGB-D cameras, Azure Kinect~\cite{kinect}, and two Kinect V2 are used. The three Azure Kinects are hard-synchronized and connected to one client desktop. Then, with three Azure Kinects, the other two Kinects are soft synchronized with the master desktop connected with a polarization camera and an event camera. Compared with expensive motion capture system used in~\cite{human36m}, our multi-camera system does not require the subject to wear a number of sensors, which gives rise to a more natural appearance in the images.

\begin{figure*}
    \centering
    \includegraphics[width=0.75\textwidth]{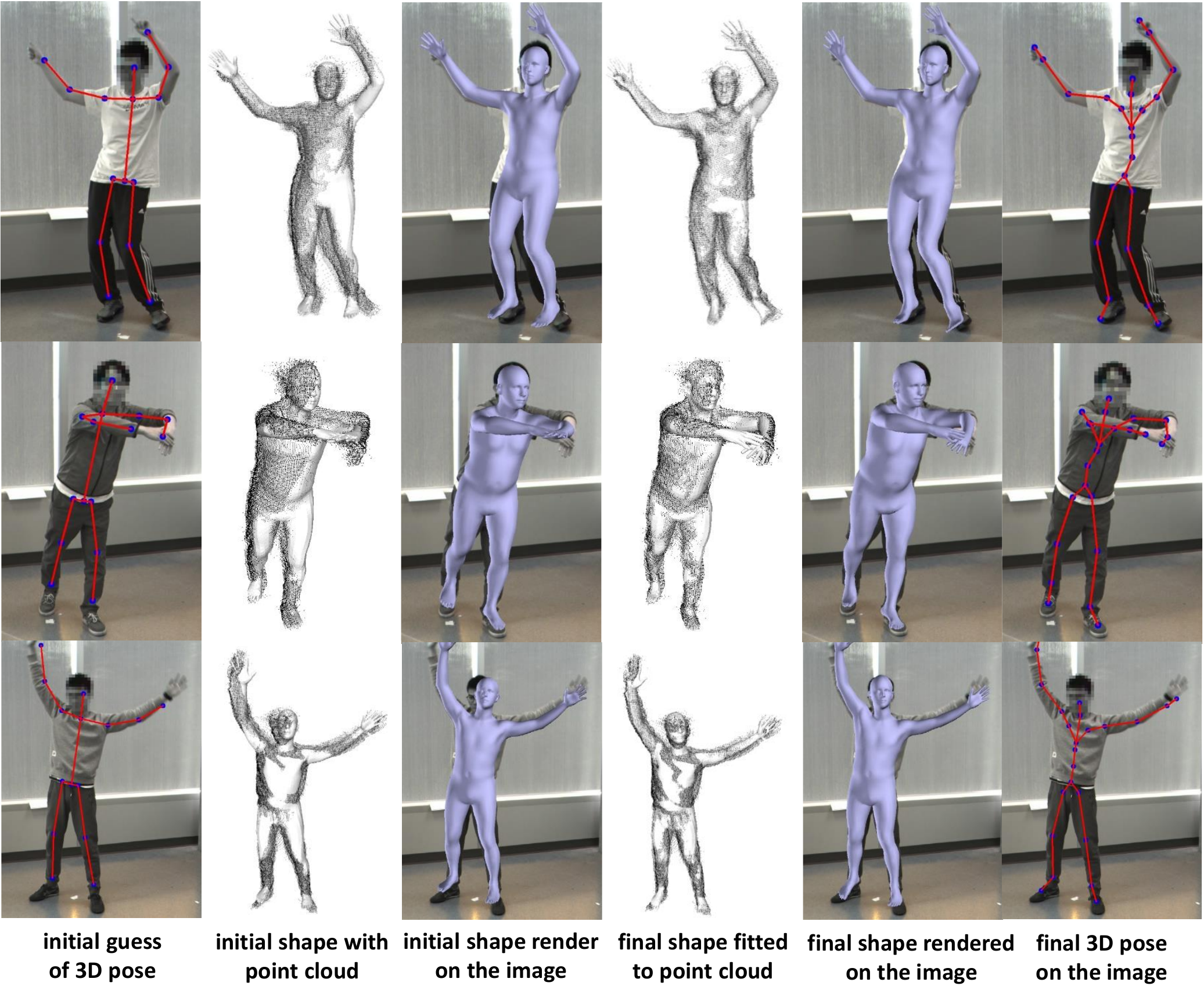}
    \caption{Exemplar figures to show that fine-tuning the initial SMPL shape to fit the point cloud can give more accurate annotated shape and pose.}
    \label{fig:ann-process}
\end{figure*}

\begin{figure*}
    \centering
    \includegraphics[width=\textwidth]{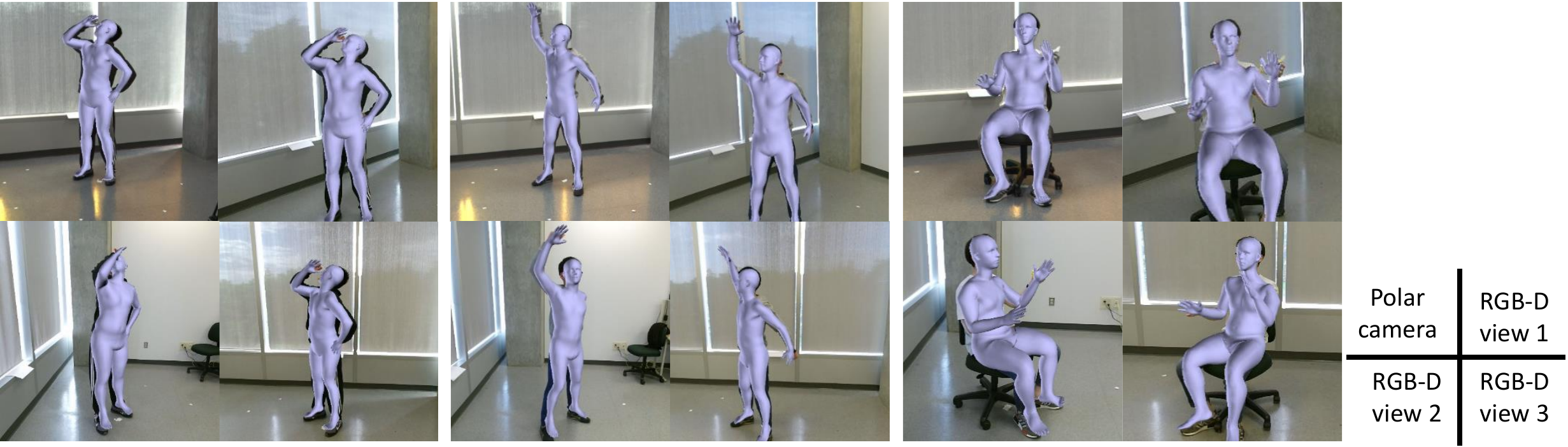}
    \caption{Exemplar multi-view figures with annotated shape and pose in PHSPDv1.}
    \label{fig:ann-dataset1}
\end{figure*}

\begin{figure*}
    \centering
    \includegraphics[width=0.85\textwidth]{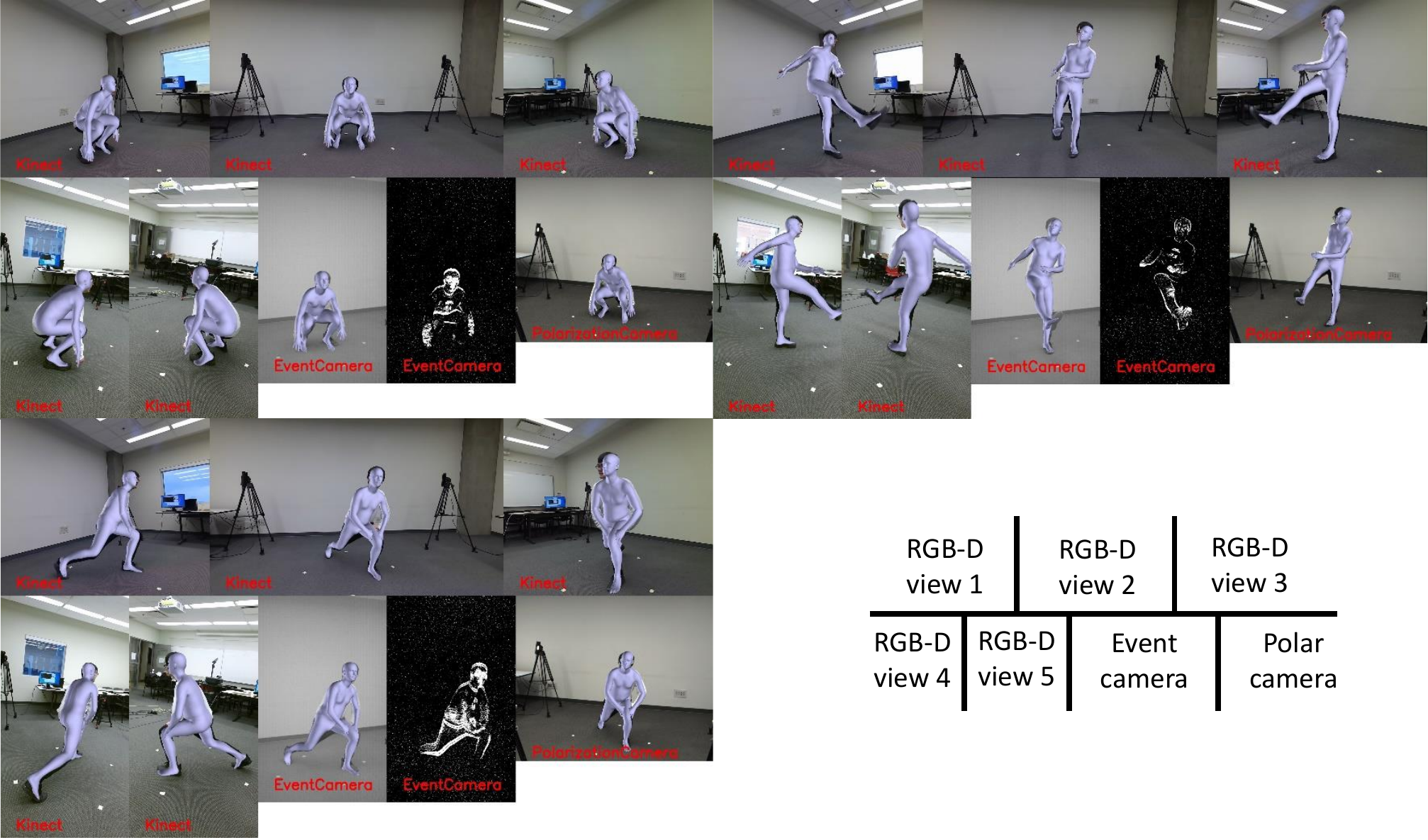}
    \caption{Exemplar multi-view figures with annotated shape and pose in PHSPDv2.}
    \label{fig:ann-dataset2}
\end{figure*}

\textbf{Annotation.} There are three main steps in the pipeline of our SMPL shape and pose annotation. (1) The first step is to obtain an initial 3D pose (joints position) from the multi-view RGB-D cameras. For each frame, the 2-D joints of all the color images are detected by OpenPose~\cite{openpose} and the depth of each 2-D joint is obtained by warping the depth image to the color image and finding the depth of its neighboring pixels. (2) The second step is to fit the SMPL male/female model to the initial pose via 3-D SMPLify-x~\cite{pavlakos2019smplifyx} and get the initial SMPL parameters. (3) The last step is to fine-tune the initial shape to fit the point cloud collected from multi-view depth images using the L-BFGS~\cite{bollapragada2018LBFGS} algorithm, where the average distance of shape vertex to its nearest point in the point cloud is minimized iteratively. Fig.~\ref{fig:ann-process} illustrates the effectiveness of the fine-tuning step to give more accurate shape and pose. Normally the initial pose is coarse because of the errors of depth or 2-D joints detection. The iterative fine-tunning process can adjust the SMPL shape to fit the human body point cloud better. Empirically, we find that SMPL male model gives better annotations than female SMPL model for female subjects in light that our recruited female subjects have similar body proportion with SMPL male model. Thus our dataset adopts SMPL male model for all subjects. Some examples of multi-view annotated shape are displayed in Fig.~\ref{fig:ann-dataset1} and~\ref{fig:ann-dataset2}.

\begin{table}
    \centering
    \begin{tabular}{|p{20pt}|p{140pt}|}
    \bottomrule \hline
        \makecell[c]{group} & \makecell[c]{actions} \\
    \hline
        \makecell[c]{1} &  \makecell[c]{warming-up, walking, running, jumping,\\ drinking, lifting dumbbells}\\
    \hline
        \makecell[c]{2} &  \makecell[c]{sitting, eating, driving, reading, \\phoning, waiting}\\
    \hline
        \makecell[c]{3} &  \makecell[c]{presenting, boxing, posing, throwing,\\ greeting, hugging, shaking hands}\\
    \hline \toprule 
    \end{tabular}
    \caption{Summary of action types performed by subjects in PHSPDv1. When recording the data, the subject is required to perform the actions within each group in random order.}
    \label{tab:dataset1-action}
\end{table}

\begin{table}
    \centering
    \begin{tabular}{|p{20pt}|p{30pt}|p{160pt}|}
    \bottomrule \hline
        \makecell[c]{group} & \makecell[c]{speed} & \makecell[c]{actions} \\
    \hline
        \makecell[c]{1} & \makecell[c]{medium} & \makecell[c]{jumping, jogging, waving hands, \\kicking legs, walk}\\
    \hline
        \makecell[c]{2} & \makecell[c]{fast} & \makecell[c]{boxing, javelin, fast running, \\ shooting basketball, kicking football, \\playing tennis, playing badminton}\\
    \hline
        \makecell[c]{3} & \makecell[c]{slow} & \makecell[c]{warming up elbow/wrist ankle/pectoral, \\lifting down-bell, squating down, drinking water}\\
    \hline \toprule 
    \end{tabular}
    \caption{Summary of action types performed by subjects in PHSPDv2. When recording the data, the subject is required to perform the actions within each group in random order.}
    \label{tab:dataset2-action}
\end{table}

\textbf{Details of PHSPD.} 
PHSPDv1 has 12 subjects, 9 male and 3 female subjects. Each subject is required to do 3 different groups of actions (18 different actions in total) for 4 times plus one free-style group. Details of actions are shown in Tab.~\ref{tab:dataset1-action}. So each subject has 13 short videos and the total number of frames for each subject is around 22K. Overall, our dataset has 287K frames with each frame including one polarization image, three color and three depth images. Quantitative details of our dataset are shown in Tab.~\ref{tab:dataset1-details} 

In PHSPDv2, 15 subjects are recruited for the data acquisition, where 11 are male and 4 are female. Each subject is required to perform 3 groups of actions (21 different actions in total, as is shown in Tab.~\ref{tab:dataset2-action}) for 4 times, where each group includes actions of fast/medium/slow speed respectively. Finally, we collect 12 short videos for each subject and each video has around 1,300 frames with 15 FPS, resulting in 178 videos in total with each video lasting about 1.5 minutes. We conduct the annotations for each video and check manually whether the annotated shape aligns well with multi-view images. We abandon the unsatisfactory annotated shapes. Details on the number of frames per subject and number of annotated frames per subject are presented in Tab.~\ref{tab:dataset2-details}. The average number of events for the dataset is around 1 million per second. Overall, our dataset consists of 240k frames with each frame including a gray-scale image and inter-frame events, a polarization image, five-view color and depth images.

\begin{table*}
    \centering
    \begin{tabular}{|c|cccc|}
    \bottomrule \hline
        \makecell[c]{subject \#} & \ gender\  & \makecell[c]{raw \#} & \makecell[c]{annotated \#} & \makecell[c]{discarded \#} \\
    \hline
        1 & female & 22561 & 22241 & 320 (1.4\%) \\
        2 & male & 24325 & 24186 & 139 (0.5\%) \\
        3 & male & 23918 & 23470 & 448 (1.8\%)\\
        4 & male & 24242 & 23906 & 336 (1.4\%)\\
        5 & male & 24823 & 23430 & 1393 (5.6\%)\\
        6 & male & 24032 & 23523 & 509 (2.1\%)\\
        7 & female & 22598 & 22362 & 236 (1.0\%)\\
        8 & male & 23965 & 23459 & 506 (2.1\%)\\
        9 & male & 24712 & 24556 & 156 (0.6\%)\\
        10 & female & 24040 & 23581 & 459 (1.9\%) \\
        11 & male & 24303 & 23795 & 508 (2.1\%)\\
        12 & male & 24355 & 23603 & 752 (3.1\%)\\
    \hline
        total & - & 287874 & 282112 & 5762 (2.0\%)\\
    \hline \toprule 
    \end{tabular}
    \caption{Detail number of frames for each subject in PHSPDv1 and also the number of frames that provides reliable SMPL shape and pose annotations.}
    \label{tab:dataset1-details}
\end{table*}

\begin{table*}[h]
    \centering
    \begin{tabular}{|c|cccc|}
    \bottomrule \hline
        \makecell[c]{subject \#} & \ gender\  & \makecell[c]{raw \#} & \makecell[c]{annotated \#} & \makecell[c]{discarded \#} \\
    \hline
        1 & male & 15911 & 15911 & 0 (0.0\%) \\
        2 & male & 15803 & 15803 & 0 (0.0\%) \\
        3 & male & 16071 & 16071 & 0 (0.0\%)\\
        4 & male & 16168 & 16152 & 16 (0.01\%)\\
        5 & male & 16278 & 16262 & 16 (0.01\%)\\
        6 & male & 16715 & 16384 & 331 (2.0\%)\\
        7 & female & 16091 & 16091 & 0 (0.0\%)\\
        8 & male & 16257 & 15642 & 715 (4.4\%)\\
        9 & male & 15467 & 15461 & 6 (0.03\%)\\
        10 & male & 16655 & 16655 & 0 (0.0\%) \\
        11 & male & 16464 & 16443 & 21 (0.13\%)\\
        12 & male & 16186 & 16186 & 0 (0.0\%)\\
        13 & female & 16064 & 14562 & 1502 (9.4\%)\\
        14 & female & 15726 & 15166 & 560 (3.6\%)\\
        15 & female & 14193 & 14075 & 118 (0.8\%)\\
    \hline
        total & - & 240049 & 236764 & 3285 (1.4\%)\\
    \hline \toprule
    \end{tabular}
    \caption{Detail number of frames for each subject in PHSPDv2 and also the number of frames that provides reliable SMPL shape and pose annotations.}
    \label{tab:dataset2-details}
\end{table*}

\bibliographystyle{IEEEtran}
\bibliography{IEEEabrv}